**Exploring Theory-Laden Observations in the Brain Basis of Emotional Experience**

Christiana Westlin[1,2,3,4], Ashutosh Singh[1,5], Deniz Erdogmus[5], Georgios Stratis[5] and Lisa Feldman Barrett[3,4,6]

[1] Indicates shared first-authorship

[2] Department of Neurology, Massachusetts General Hospital, Harvard Medical School, Boston, MA 02115

[3] Department of Psychiatry, Massachusetts General Hospital, Harvard Medical School, Boston, MA 02115

[4] A. A. Martinos Center for Biomedical Imaging, Massachusetts General Hospital, Harvard Medical School, Boston, MA 02115

[5] Department of Electrical & Computer Engineering, Northeastern University, Boston, MA 02115

[6] Department of Psychology, Northeastern University, Boston, MA 02115

\***Corresponding author:**
Christiana Westlin, PhD
Massachusetts General Hospital
55 Fruit Street, Boston, MA, USA
(978)-618-9876
cwestlin@mgh.harvard.edu

**Abstract**
In the science of emotion, it is widely assumed that folk emotion categories form a biological and psychological typology, and studies are routinely designed and analyzed to identify emotion-specific patterns. This approach shapes the observations that studies report, ultimately reinforcing the assumption that guided the investigation. Here, we reanalyzed data from one such typologically-guided study that reported mappings between individual brain patterns and group-averaged ratings of 34 emotion categories. Our reanalysis was guided by an alternative view of emotion categories as populations of variable, situated instances, and which predicts a priori that there will be significant variation in brain patterns within a category across instances. Correspondingly, our analysis made minimal assumptions about the structure of the variance present in the data. As predicted, we did not observe the original mappings and instead observed significant variation across individuals. These findings demonstrate how starting assumptions can ultimately impact scientific conclusions and suggest that a hypothesis must be supported using multiple analytic methods before it is taken seriously.

**Significance Statement**

When studying the brain basis of emotion, scientists often use methods that are guided by certain theoretical assumptions. This approach constrains what is possible for researchers to observe, and ultimately results in findings that reinforce the assumptions that guided the investigation in the first place (also called *theory-laden* observations). In contrast, discovery-based analyses do not constrain what is possible to observe. In this study, we contrasted a previously published confirmatory-based analysis with a discovery-based re-analysis of the same data, to demonstrate how two different analyses of the same data, guided by different starting assumptions, can lead to very different conclusions. These findings highlight the value of discovery-based methods and the importance of carefully considering assumptions in scientific research.

Initial assumptions about the nature of psychological phenomena influence study design and analysis, thereby constraining what can be observed and learned from those studies (Barrett, 2022; Dubova & Goldstone, 2023; Westlin et al., 2023). Emotion research is a particularly good example of the ways in which starting assumptions constrain subsequent knowledge generation. Most published neuroimaging studies of emotion are designed to test the hypothesis that certain Western folk categories of emotion (e.g., anger, sadness, fear, happiness, etc.) are biological and psychological types, each with its own diagnostic prototype. Participants are routinely exposed to a relatively small sample of stimuli that are assumed to evoke typical instances of each folk category (e.g., a set of video clips) while they undergo brain scanning. The goal is to search for one single spatial pattern of brain activity (measured via the blood-oxygen level dependent (BOLD) response) that uniquely corresponds to each folk category and that is robust across individuals and situations (e.g., Camacho et al., 2023; Kassam et al., 2013; Kragel & LaBar, 2015; Saarimäki et al., 2016, 2018; Wager et al., 2015; Zhou et al., 2021). Correspondingly, studies group data across all trials for a given category based on either experimenter-assigned labels or on group-averaged ratings of emotion (often collected from independent samples using a restricted list of categories), and then try to identify a relationship between group-averaged BOLD response patterns and these emotion category labels or ratings, to ultimately provide evidence in support of the emotion typology that inspired the research in the first place.

Studies designed in this way – failing to sample the full range of variation within each category – are not optimally equipped to observe meaningful within-category variation that is structured across contexts and individuals. Such structured variation is hypothesized, a priori, as part of a relational theory of emotion, which hypothesized that instances of emotion are intrinsically contextual and situated (Barrett, 2017b, 2022; Barrett & Lida, 2024; Barrett & Theriault, in press). A growing number of studies have observed reliable, structured within-category variation, and/or across-category similarity, in the magnitude of the BOLD response (e.g., Caseras et al., 2010; Lebois et al., 2020; Schienle et al., 2013; Wilson-Mendenhall et al., 2011, 2015, 2019), in dynamics of the BOLD response (e.g., Chang et al., 2021; Singh et al., 2021), and in functional connectivity (e.g., Doyle et al., 2022; Raz et al., 2016). The hypothesis of contextually-structured variation within a biological category is consistent with Darwin's idea that a species is a category of highly variable individuals (Darwin, 1859), called population thinking (Mayr, 1994). In population thinking, structured variation within a biological category is real and statistical aggregates that summarize the category's distribution are a simplification that is a fiction (similar to "the evils of averaging" (Estes, 1956; Gallistel, 2012), in which the statistical average for a sample of participants is rarely, if ever, observed in any individual members of that sample; for an example pertaining to emotions using mathematical simulations, see Clark-Polner et al., 2017).

To allow for the possibility of observing meaningful within-category variation, one avenue for improvement on the standard experimental design would be to sample a larger variety of evocative stimuli. A recent study on the brain basis of emotion did just that, scanning the brains of five participants with functional magnetic resonance imaging (fMRI) as they watched over 2000 video clips intended to induce instances of 34 folk emotion categories (Horikawa et al., 2020). These clips had been labeled by separate subsamples of individuals (N = 9 to 17 in each subsample) who reported their emotional experiences using 34 emotion words (e.g., anger, fear, happiness, etc.) as they watched subsets of 30 film clips. The words were chosen by the experimenters to match the emotion categories of movies they sampled. Ratings were averaged across all raters to produce mean ratings across 34 emotion labels for each clip (Cowen & Keltner, 2017). Each video was also rated by separate samples of participants across 14 affective/appraisal features, and 73 semantic features, which were similarly group-averaged across all raters. Although the study's design was better suited than prior studies to detect variation across a large number of sampled instances, the analyses reported in Horikawa et al. were designed to observe folk emotion types. Specifically, the authors conducted three primary analyses of the data: (1) predicting group-averaged emotion category ratings from each independent participant's brain data, (2) predicting each participant's brain data from group-averaged emotion category ratings, and (3) relating data-driven clusters of each independent participant's brain data to the group-averaged emotion category ratings.

In this paper, we specifically focus on the third, unsupervised clustering analysis, to demonstrate how the conclusions drawn from such analyses are highly contingent on underlying modeling assumptions. We have previously shown this to be the case with the movie rating data from Cowen & Keltner (2017), calling into question their conclusions (Azari et al., 2020). We note that the authors also conducted analyses using group-averaged affective and appraisal feature ratings, although these ratings share many of the same issues we describe below. Because the authors reported a stronger relationship between the neural data and emotion category ratings, we limit our focus to analyses conducted using the group-averaged emotion category ratings. We describe below several of the methodological decisions in Horikawa et al. that reflect an underlying assumption that the BOLD signals measured during emotional instances could be organized according to a folk emotion typology:

**Use of Binarized, Group-Averaged Ratings**

To derive meaning from brain signals measured as five participants viewed a series of video stimuli, Horikawa and colleagues used ratings for each corresponding video that were reported in a previously published study (Cowen & Keltner, 2017). In this study, an independent sample of participants rated the extent to which a given video made them feel each of 34 different emotion words. The ratings were made on a scale from 0-100, and then converted to binarized {0,1} ratings such that any score greater than zero was assigned a one. Binarization assumes similarity across all instances of a particular emotion category, irrespective of the intensity of each experience, and consequently constrains the ability to observe variation across participants. Horikawa and colleagues used continuous ratings in supplementary analyses, but all analyses reported in the main text involved binarization. In the main analyses, the binarized ratings were averaged across the independent sample of raters. Aggregating across participants assumes that the group-averaged statistic is representative of individual experience (a key typological assumption), and thus further masks any potentially meaningful variation. By relying on group-averaged, binarized ratings to make sense of the brain data of each independent participant, the analyses reported by Horikawa and colleagues did not allow for the possibility of variable, participant-specific emotion categories. In other words, the analyses were set up to provide evidence in support of the normative categories, thereby limiting what was possible to observe.

**Feature Selection**

Feature selection was a necessary step prior to clustering due to the computational demands of analyzing high-dimensional data. Specifically, Howikawa et al. reduced the dimensionality of the whole-brain BOLD data by selecting a subset of voxels that were best predicted by the group-averaged ratings in a prior supervised analysis. The validity of this practice has been the subject of vigorous debate (see Vul et al., 2009 and commentaries), but our point here is that this procedure reduced the possibility of observing BOLD signal patterns that might be rare but meaningful. The features input into the clustering analysis were therefore only voxels whose signals were most related to the group-averaged ratings, which predisposed the clustering analysis to provide evidence in support of a normative structure. This consequently limited the ability to observe structure that meaningfully varied from the assumed normative structure.

**Pre-Specifying the Number of Clusters**

K-means clustering was used to estimate structure in the BOLD data of each individual participant. Horikawa et al. did not *empirically* estimate the best-fit number of clusters for each individual (e.g., through model-order selection using Bayesian Information Criterion (Schwarz, 1978), as in Le Mau et al., 2021). Instead, they pre-specified the same number of clusters for all participants, K = 27 (as well as K = 15 or 50 in supplemental analyses) based on group-level findings reported in Cowen & Keltner (2017). Imposing the same number of clusters across all participants makes the assumption that the brain signals measured while viewing emotional stimuli were organized with the same underlying structure across participants. This approach further limits any ability to observe variable structure across participants. Evidence from unsupervised analyses of BOLD data (Doyle et al., 2022; Singh et al., 2021) and peripheral physiological signals during emotion (Hoemann et al., 2020) show clearly that meaningful cluster structure varies across individuals who are performing the same task or experiencing instances of the same emotion category.

These analytic decisions largely constrained what was possible to observe in Horikawa et al.'s clustering analysis because they removed the possibility of observing meaningful, structured variation across individuals. As a consequence, typological hypotheses were compared only to the null hypothesis (no structure) rather than to the meaningful, alternative hypothesis of within-category, structured variation. Rejecting the null hypothesis led to a conclusion of de facto support for the typology theory of emotions, reinforcing the assumptions that constrained their investigation in the first place. This is the very definition of theory-laden experimental design (Dubova & Goldstone, 2023).

In the present study, we re-analyzed the Horikawa et al. data with the goal of demonstrating how assumptions about the phenomena under investigation can profoundly impact the conclusions that are possible to draw from that investigation. We used a discovery-based approach that made only basic modeling assumptions (which we explicitly acknowledged up-front), and that didn't constrain how the dominant sources of variance in the BOLD signals would be structured for each participant. Specifically, we modeled whole-brain patterns of BOLD signal across all trials for a given participant by first reducing the dimensionality of each person's data separately, using principal component analysis (PCA; Jolliffe & Cadima, 2016). We then used Gaussian Mixture Modeling (Bishop, 2006) to cluster the data for each participant. We ensured that the clusters were reliable and therefore reflected meaningful, structured variance by examining cluster stability across multiple reinitializations of the model. This discovery-based approach allowed us to examine how the variance across each participant's whole-brain BOLD signal was structured during each instance of emotion, removing any typological constraints on this structure. If this approach revealed clusters[1] of BOLD data that are consistent across individuals, with structure that is highly related to the group-averaged emotion category ratings, then these findings, consistent with those of the original analysis, would support the conclusions offered in the Horikawa et al.. If, however, our re-analysis revealed little correspondence between the number of BOLD signal clusters across participants, as well as little correspondence between clusters and the group-averaged emotion category ratings, then it would demonstrate how different analytical approaches, guided by different underlying assumptions, can lead to divergent conclusions, emphasizing the need for researchers to routinely consider the theory-laden assumptions that are embedded in their design and analytic choices.

---

[1] Here we use the term 'clusters' to refer to the number of Gaussian components, although we note that strictly speaking, the number of clusters produced by a Gaussian mixture model may be less or more than the number of Gaussian components.

## Results
**Estimating Participant-Specific Clusters**

We first treated the number of clusters of BOLD signal data as a parameter to be learned and estimated clusters separately for each participant. We reduced the dimensionality of each participant's data using PCA, followed by Gaussian Mixture Modeling (Bishop, 2006) of the lower-dimensional data for all trials for a given participant. We used BIC (Schwarz, 1978), a model-order selection method, to select the optimal number of components (i.e., clusters) for each participant's Gaussian Mixture Model (GMM). Plots of BIC revealed variable numbers of clusters that best fit the data across participants, ranging from 11 to 14 (Figure 1). Such variation in the number of empirically-estimated clusters indicated that there was unique structure present in each participants' BOLD data as every participant watched the same 2196 video clips, each intended to induce the same emotional experience in everyone who watched it. For a comparison of cluster means across participants, see Supplementary Materials.

The variation in cluster structure across participants was not due to a lack of clustering stability; the average Rand index (Rand, 1971) across 10 re-initializations of each participant's model was 0.85 (0 = low stability, or different clusterings every time a model was initialized for a participant's data; 1 = high stability, or exactly the same clusterings for every initialization), indicating that trials clustered together in highly similar ways across multiple initializations of each participant's model. This finding indicates that the variation in cluster structure across participants was reliable.

The observed variation across clusters was also not due to the model's inability to recover true categories in the data (we previously validated the model's ability to estimate biological categories if they are present in the data, such that the model accurately estimated the true number of clusters in synthetic BOLD signal data; Azari et al., 2020). The variation was also not due to problems occurring with PCA in the case of high-dimensional data with a low sample size (see Supplemental Materials for a thorough test of problems in principal component estimations).

**Relating Cluster Structure to Assumed Emotion Types**

Each datapoint (i.e., lower dimensional BOLD data corresponding to a single video-watching trial) in a given participant's model had some probability of belonging to each cluster. In our analysis, each datapoint had a very high probability (1 or nearly 1) of belonging to only a single cluster, and a very low probability (0 or nearly 0) of belonging to all other clusters, indicating that the clusters were extremely separable. We therefore were able to consider only one cluster assignment per trial to investigate each cluster based on its most probable trials.

*Trial Structure*

If emotion categories are organized as types, then we would expect participants to have a highly similar experience as they viewed the same video, as intended by the study design. We therefore examined whether the trials during which participants viewed the same videos clustered in similar ways across people. To do this, we computed the percent of trials that overlapped between each pair of clusters across all participants (Figure 2). The percent overlap for all comparisons within a participant was zero, since each trial was assigned to only one cluster. The mean percent overlap for all comparisons across participants was 7.3096% (SD = 7.8075). In other words, very few trials overlapped between clusters, indicating that trials clustered in unique ways across people, despite participants watching the same video clips during these trials, each intended to induce the same instance of emotion in everyone who watched it. Participants also differed in terms of their clustering similarity to other participants. For example, four clusters for Participant 1 had greater than 40% overlapping trials with two clusters each from Participants 2 and 4, whereas Participants 3 and 5 had no clusters that reached a threshold of 40% overlap. Plots of percent overlap at varying thresholds across all comparisons are shown in Supplementary Figure 2.

*Emotion Category Ratings*

We next compared the participant-level clusters of BOLD data to group-averaged emotion category ratings through both qualitative and quantitative analyses. Both analyses revealed that the empirically estimated clusters based on similarities in BOLD signal data did not cleanly map to the subjective ratings made by another sample of participants when watching the same clips.

Using normalized mutual information (NMI), we examined the amount of information shared between participant clusterings and emotion category ratings. A NMI value of zero would indicate complete independence between the two variables (i.e., no information is shared between a given rating and the clustering), while a NMI value of one would indicate complete dependence (i.e., all information is shared between a given rating and the clustering). All NMI values computed between each rating and participant clusterings were less than 0.01 (M = 0.0141, SD = 0.0135), indicating that the individual-level clusterings share almost no information with the group-averaged emotion category ratings that were obtained from an independent sample of participants (Figure 3).

We then qualitatively examined the relationship between ratings and clusterings. If a given folk emotion category mapped singularly to a cluster of BOLD data, then we would expect that cluster to be comprised only of trials that were rated highly for that emotion category. Thus, for data visualization purposes, we assigned only a single label to each trial based on the highest rated category, and then examined the distribution of labels within each cluster (note: this assumption was made for data visualization purposes only, and incorrectly ignores other categories that were also endorsed for a given trial). We observed a mixture of trials labeled by each emotion category across each estimated cluster, and across all participants (see Figure 4a for an example from one representative participant; plots of all participants are shown in Supplementary Figure 3).

**Relating Cluster Structure to Other Known Sources of Variation**

Finally, we explored several other potential sources of variation beyond emotion categories to investigate whether the variation in cluster structure reflected any of these sources. Each of these sources of variation was conducted using the same qualitative (mutual information) and quantitative (visualization of top-rated trials) analyses as was used for emotion category ratings.

*Affective and Appraisal Feature Ratings*

We first examined the relationship between cluster structure and a set of affective and appraisal feature ratings averaged across an independent sample of participants (Cowen & Keltner, 2017). Similar to the findings from emotion category ratings, NMI values between clusters and feature ratings were all less than 0.005 (M = 0.0019, SD = 0.0058; Supplementary Figure 4a), indicating that the individual-level clusterings and group-averaged affective and appraisal ratings shared almost no information. We also observed a mixture of features across each cluster (Figure 4b; Supplementary Figure 5).

*Semantic Feature Ratings*

We next examined a set of 73 semantic feature ratings made by an independent sample of participants (Horikawa et al., 2020). NMI values between clusters and semantic features were again close to zero; mean NMI across participants was 0.0260 (SD = 0.0292; Supplementary Figure 4b). When examining the top-rated features for trials in each cluster, we again observed a heterogeneous distribution of semantic ratings across clusters (Figure 4c; Supplementary Figure 6).

*Run and Session Ordering*

Finally, we examined whether the cluster structure for each participant was related to the scanning session or run of corresponding trials in each cluster (Singh et al., 2022). Mean NMI values between the clusters and scanning sessions/runs were again quite low (session: M =

0.1882, SD = 0.0269; run: M = 0.1720, SD = 0.0134; Supplementary Figure 4c). Visualizations of the sessions and runs for which the most probable trials in each cluster belonged also revealed a heterogeneous mixture of trials for each session and run across each cluster (Figure 4d,e; Supplementary Figure 7).

## Discussion

This re-analysis of published data demonstrates the impact that starting assumptions have on study design and analyses, influencing what is possible to observe and learn from an investigation. To demonstrate this, we re-analyzed data from a recent study (Horikawa et al., 2020) that reported evidence in support of a typological theory of emotion. Horikawa et al. reported that clusters of each individual participant's BOLD data (measured as they viewed a series of emotionally evocative videos) could be explained by normative emotion category ratings (averaged from an independent sample of individuals who viewed the same videos). The typological assumptions implicit in several of their methodological decisions (including the use of group-averaged ratings, analyzing voxels that were most related to these ratings, and pre-specifying the same number of clusters across participants) substantially impacted their results and their inferences about the meaning of those results.

The results of the Gaussian Mixture Modeling re-analysis indicated that the variance within each person's BOLD signal data was reliably structured in a unique way, despite their watching the same video clips that were selected to induce the same emotional instances. This variable structure across participants is concealed in analyses that are not suited to detect individual variation, such as confirmatory, group-based analyses. Additionally, the number of clusters estimated for any given participant (11 to 14) was not the same as the number of clusters assumed in the primary clustering analyses by Horikawa et al. (k = 27).

Furthermore, our model-first, discovery-based re-analysis did not lead to the same conclusion of predictable mappings between individual BOLD signal patterns and normative emotion category ratings, but rather provided evidence of structured variation across participants. Each unique clustering structure shared little to no correspondence with the group-averaged ratings. It is likely that individual variation in the emotion category ratings is being masked due to group-averaging. A prior re-analysis of the normative rating data using unsupervised clustering also did not replicate the 27 categories previously reported for this data (Azari et al., 2020).

If we assume that the methods applied in both the present study, and in the original analysis, were executed correctly, then the different findings between the two investigations lies in the assumptions and modeling decisions. Several analytical decisions in the original analysis (outlined in the introduction) were rooted in a typological view of the mind that has dominated psychological research since the 19th century, and continues to guide experimental design and data analysis in much of the research that is published in the present day (Lewontin, 2000; Mayr, 1994). A typological view of emotion assumes that emotion categories are natural kind categories (i.e., a grouping of instances with certain prototypic features that they share; for a discussion, see Barrett, 2022; Barrett & Theriault, in press). Individual instances may vary slightly in their features, and instances at the boundaries of a category may share features with instances of other categories, but the majority of instances are assumed to fit firmly within the category under a typological view. This view is central to the recently proposed semantic space theory, which hypothesizes that upwards of 20 distinct categories of emotion organize emotional experience and expression (Cowen & Keltner, 2021; Keltner et al., 2023). Horikawa and colleagues made several analytical decisions that were influenced by semantic space theory (and thus, a typological view), and that ultimately confirmed the assumption of the categorical structure proposed by semantic space theory.

In contrast, our analysis was guided by a relational view of emotion as contextual and situated (Barrett, 2022; Barrett & Lida, 2024; Barrett & Theriault, in press), which a priori hypothesizes that folk emotion categories are populations of highly variable instances (Barrett, 2013, 2017a, 2017b; Clark-Polner et al., 2017; Siegel et al., 2018). To properly test this hypothesis, an analytic approach must allow for the possibility of observing meaningfully structured variation, yet not prevent the ability to observe a typological structure if one did indeed exist. Although we used less constrained, more objective methods for both feature selection and model-order selection, our analysis, like any model-based approach, is only as valid as our basic modeling assumptions.

Future studies should thus investigate the neural basis of emotional experience via alternative models. Most typologically-inspired studies compare their hypotheses to the null hypothesis rather than design their experiments to consider alternative hypotheses in a substantive manner. Future work should seek to systematically compare one theory's hypotheses to another's rather than to the null. Additionally, the use of multiple methods and modeling assumptions in a single study, such as a multiverse analysis, can provide valuable insight into how methodological decisions can impact modeling solutions (e.g., Azari et al., 2020; Le Mau et al., 2021).

Our analyses also revealed variation in the similarity across participants (e.g., one participant had four clusters that shared >40% of the same trials with other participants' clusters, whereas two participants had no clusters that met this threshold). Future studies investigating such variation across a larger number of participants are needed to understand factors that might be structuring similarity across participants. For example, it is possible that demographic or life history variables might predict the structure of who experiences a given movie clip in the same way, or that clips might be experienced similarly if they contain certain relational, culture, or other themes beyond those captured by the available semantic ratings.

Finally, the mutual information between averaged ratings of affective/appraisal features or semantic features of the corresponding video stimuli in each cluster and each participant's clustering was quite low, suggesting that cluster structures was independent of any of the sources of variation that were available to be tested. If we assume that our basic modeling assumptions are valid (which we previously tested in Azari et al., 2020, and which we expand on in the present study through a test of our principal component estimations), then our conclusions are not invalidated by an inability to identify the features structuring the variance. It remains possible that the available psychological features could still explain variation in BOLD data, yet these ratings were averaged across different individuals than those whose brains were scanned, and so the same concerns hold for these ratings as for the emotion category ratings. Specifically, a main limitation of the Horikawa et al. data set was that it did not contain individual-level features that may have been structuring the observed variance. To fully explore the structure in these data-driven clusters, future research is needed in which participants are not only densely sampled as their brains are scanned, but also as ratings of their self-reported experiences are collected. It is also possible that the features are not mutually independent from each other, suggesting that their interactions should be modeled as a nonlinear function over these features to explain potential variation seen in the BOLD signal (e.g., Kwon & Zou, 2022; Masoomi et al., 2023). It is also possible that the cluster structure cannot be explained by the presently available features, but that is a topic of future research.

## Conclusions

This report illustrates some of the ways in which starting assumptions can guide design and analytical decisions, resulting in very different conclusions. In the study of emotion, studies often take a confirmatory-based approach: folk emotion categories are assumed to form a biological and psychological typology, and this assumption guides empirical design and analytical practices, resulting in confirmation of a typology. Yet the findings from these studies have been shaped by theoretical assumptions, and thus are theory-laden, thereby constraining the nature of variation and the type of structure they permit. In contrast, discovery-based approaches, such as the approach taken in our re-analysis, allow for the observation of meaningful, structured variance, while still preserving the ability for a typology to be observed, should it exist. In a discovery-based approach, researchers state assumptions up front, and all inference and conclusions are explicitly conditioned on such assumptions. Such an approach makes it explicitly clear what is, and what is not, possible to discover under the stated assumptions. Given the state of knowledge in the science of emotion, future research using discovery-based approaches is vital to improve our understanding of how instances of emotion are represented in the brain, as well as in the body and in behavior.

## Materials and Methods

### Dataset

We analyzed data previously reported in Horikawa et al., 2020, which includes comprehensive study design details. Briefly, five participants (1 female, ages 22 to 34) participated in the study. All participants provided written informed consent, and the study protocol was approved by the Ethics Committee of the original study's institution (Advanced Telecommunications Research Institute International). All methods were carried out in accordance with relevant guidelines and regulations. Participants completed 61 runs across eight fMRI scanning sessions, during which they freely viewed 2196 short, silent videos (0.15 to 90s) from (Cowen & Keltner, 2017) that were selected to induce instances of 34 emotion categories. Each scanning run consisted of 36 blocks, during which a video was presented followed by a 2s rest period. Videos shorter than 8s were repeatedly presented for 8s, those longer than 8s were presented once in their entirety.

### *Feature Ratings*

Each video clip was previously rated across 34 emotion category labels and 14 affective/appraisal features, as described in (Cowen & Keltner, 2017). Ratings for 73 semantic features were also collected by Horikawa et al. (2020). We used each of these available feature ratings to examine whether the participant-specific clusters of brain data shared any variance with the ratings. Details of each rating type are described in corresponding sub-sections below.

**Emotion Category Ratings.** Each video clip was previously rated by separate subsamples of 9 to 17 raters using 34 emotion category labels. Participants rated each video according to the degree to which it made them feel each of the following 34 emotion categories along a 100-point scale: admiration, adoration, aesthetic appreciation, amusement, anger, anxiety, awe, awkwardness, boredom, calmness, confusion, contempt, craving, disappointment, disgust, empathic pain, entrancement, envy, excitement, fear, guilt, horror, interest, joy, nostalgia, pride, relief, romance, sadness, satisfaction, sexual desire, surprise, sympathy, triumph. In the original study that collected these ratings (Cowen & Keltner, 2017), these ratings were binarized on a {0,1} scale, followed by group-averaging of the binary scores across all raters, to provide a single value for each of the 34 categories per video. In their primary analyses, Horikawa et al. used these binarized, group-averaged ratings. They also reported supplementary analyses using scores that were averaged across the 100-point ratings without binarization. We utilized the non-binarized ratings in our analyses, which consisted of a single group-averaged rating between 0 and 100 for each of the 34 categories per video.

**Affective/Appraisal Feature Ratings.** Different subsamples of raters also rated their experience of each video (N = 9 per video) using 14 affective/appraisal features: approach, arousal, attention, certainty, commitment, control, dominance, effort, fairness, identity, obstruction, safety, upswing, valence. Each rating was made on a Likert scale from 1 to 9, with 5 anchored at neutral (see Cowen & Keltner, 2017 for more detail). The ratings were averaged across all raters, resulting in 14 group-averaged affective/appraisal feature ratings per video.

**Semantic Feature Ratings.** A third sample of raters previously rated each video (N = 12 per video) along a set of 73 semantic features using a dichotomous yes/no response as to whether each feature was present in the video. The semantic features included objects, scenes, actions, and events (see Horikawa et al. (2020) for a full description). The ratings were averaged across all raters, resulting in 73 group-averaged semantic feature ratings per video.

**BOLD Data Acquisition and Preprocessing**

MRI data were collected using a Siemens MAGNETOM Verio 3T scanner. A full description of the MRI acquisition parameters can be found in (Horikawa et al., 2020). We performed data preprocessing using fMRIprep 20.0.6 (Esteban et al., 2019), which is based on Nipype 1.4.2 (Gorgolewski et al., 2011, 2017). For each of the 61 runs per participant, we performed the following preprocessing steps on the raw data obtained from Horikawa et al.. First, a reference volume and its skull-stripped version were generated using a custom methodology of fMRIPrep. A deformation field to correct for susceptibility distortions was estimated based on fMRIPrep's fieldmap-less approach, and a corrected EPI (echo-planar imaging) reference was calculated based on the estimated susceptibility distortion for a more accurate co-registration with the anatomical reference. The BOLD reference was then co-registered to the T1w reference using flirt (FSL 5.0.9; Jenkinson & Smith, 2001) with the boundary-based registration (Greve & Fischl, 2009) cost-function. Head-motion parameters with respect to the BOLD reference (transformation matrices, and six corresponding rotation and translation parameters) are estimated before any spatiotemporal filtering using mcflirt (FSL 5.0.9; Jenkinson et al., 2002). BOLD runs were slice-time corrected using 3dTshift from AFNI 20160207 (Cox, 1996). The BOLD time-series were then resampled into standard MNI152NLin2009cAsym space. Several confounding time-series were calculated based on: framewise displacement (FD), DVARS and region-wise global signals extracted within the cerebrospinal fluid (CSF) and the white matter (WM). The confounding time series derived from head motion estimates and global signals were expanded with the inclusion of temporal derivatives and quadratic terms for each (Satterthwaite et al., 2013). Frames that exceeded a threshold of 0.5 mm FD or 1.5 standardized DVARS were annotated as motion outliers.

The following steps were performed on resulting functional images in Nilearn (https://nilearn.github.io/). Images were smoothed using a 4mm FWHM Gaussian kernel. The data were then modeled using a general linear model (GLM) with a separate regressor for each trial as well as the following confounding variables: (1) six head-motion parameters, (2) averaged cerebrospinal fluid signal, (3) averaged white matter signal, (4) the temporal derivatives and quadratic terms of (1), (2), and (3), (5) motion spikes (defined as frames exceeding 0.5mm FD or 1.5 standardized DVARS), and (6) cosine drift model regressors with a cutoff of 0.01 Hz. The GLM convolved each regressor with the canonical hemodynamic response function, and prewhitening was performed through a first-order autoregressive model to account for the temporal autocorrelations. This procedure yielded a total of 2196 brain maps per participant.

**Unsupervised Clustering**

We used a combination of dimensionality reduction via PCA (Jolliffe & Cadima, 2016) and Gaussian Mixture Modeling, as described in the following sections (and previously described in Azari et al., 2020). Both PCA and Gaussian Mixture Modeling are data-driven methods that allowed us to examine how the variance across each participant's whole-brain BOLD signal was structured during each instance of emotion they experienced. Both PCA and Gaussian Mixture Modeling carry their own basic modeling assumptions, which are described in each of the corresponding sub-sections. All analyses were conducted in Python.

***Dimensionality Reduction via PCA***

Whole-brain fMRI BOLD signals for an individual participant are very high in dimensionality because the number of voxels across the brain is extremely large. This dimensionality makes computations prohibitive, so dimensionality reduction is a common practice for fMRI analyses (Mwangi et al., 2014). We use PCA for dimensionality reduction because it is an orthogonal linear dimensionality reduction method that preserves the variance in the original data as much as possible when performed correctly (i.e., the predominant sources of variation in the higher-dimensional data will be maintained in the lower-dimensional projection). Our use of dimensionality reduction is therefore based on the assumption that fMRI BOLD data can be projected to a lower dimensional surface without losing any reliable information because some voxels of fMRI data contain mutual information (i.e., their BOLD signals are correlated).

We applied PCA using the Scikit-learn implementation to reduce the dimensionality of our high-dimensional (1082035 voxels) brain maps for each participant. We used the same number of principal components for all participants, which we selected based on the smallest number of components that explained at least 95% of the variance in each participant. This resulted in a feature vector of length 279 for each trial for each participant that was used as input into our model.

For most fMRI-based studies (including the present study), the number of original dimensions of the data is much larger than the number of samples. For such a scenario, applying PCA can result in certain phenomena like upward bias in sample eigenvalues and inconsistency of sample eigenvectors that may bias the results as discussed in (Johnstone & Paul, 2018). We performed a series of tests, as explained in the Supplementary Methods, to investigate the consistency of the most dominant eigenvectors in our data, spread of the eigenvalues and to understand the structure of the covariance matrix. These tests revealed a few dominant eigenvectors for the covariance matrix that were consistent across iterations of randomly sampling the data, suggesting that PCA is appropriate to use with our data. The orthogonal linear projection of PCA does not spatially deform data distributions and is thus particularly useful with our modeling approach that involves data probability density modeling.

### *Gaussian Mixture Modeling*

We used Gaussian Mixture Models (GMMs; Bishop, 2006) to estimate structure in each participant's data, which is consistent with our goal of modeling the structure in participant-specific patterns of brain activity during emotional instances. GMMs assume that the data are generated from a mixture of Gaussian distributions with unknown parameters (i.e., mean, covariance, and mixture coefficients) and use expectation-maximization, an iterative algorithm, to estimate these parameters, considering all features simultaneously (Bishop, 2006). GMMs have been used to cluster fMRI BOLD data across several domains of research, based on the assumption that fMRI data are Gaussian distributed (Garg et al., 2011; Roge et al., 2015; Vul et al., 2012). If high dimensional fMRI BOLD data exhibits a GMM distribution, the PCA-reduced dimensional data will also exhibit a GMM distribution. We have previously demonstrated that this model is sufficiently sensitive to detect a meaningful categorical structure in the data if one exists (Azari et al., 2020). Specifically, we generated synthetic fMRI BOLD data with three clear, separable categories and we were able to successfully recover three clusters corresponding to these three categories using this analytic method .

To estimate the optimal number of clusters to use in each model (we refer to model components as clusters), we performed model-order selection using Bayesian Information Criterion (Schwarz, 1978) with 100 initializations across a range of 0 to 30 components. BIC is an *objective* means of selecting which solution best describes the structure present in the data (i.e., independent of any theoretical or human bias, other than what is already present in the data because of the stimuli, participants, or experimental methods used). BIC selects the most likely model amongst a set of possible models by simultaneously considering the likelihood of data under each model (quality-of-fit) and the complexity of each model by including a penalty term for the number of free parameters to avoid overfitting (Schwarz, 1978). After determining the optimal number of components from BIC, the Scikit-learn implementation of GMM with full covariance was used to discover the number and membership of clusters for each participant.

**Probabilistic Graphical Model**

The probabilistic graphical model (PGM) that describes our modeling assumptions is shown in Supplementary Figure 1. Clear circles reflect random variables that are not observed, while shaded circles indicate random variables that are observed. Directed arrows represent the statistical dependence between pairs of random variables.

The inputs to the model are fMRI data $X$, where the data points are represented by the random variable $x \in R^M$, and $M$ represents the dimensions of the fMRI data. We assume that there exists a latent variable $y \in R^D$, where $D$ is the dimension of this latent space such that $D$ is less than $M$. We make this assumption because some of the dimensions (voxels) of fMRI signals may be correlated. If we assume that the distribution of *y* is Gaussian (or a mixture of Gaussians) and the conditional distribution of $x|y$ is also Gaussian (or a mixture of Gaussians), then the first $D$ principal components from PCA give the maximum likelihood solution. Under these assumptions the marginal distribution $p(x)$ will also be Gaussian. We utilized PCA for dimensionality reduction because it is an orthogonal linear projection, such that a GMM for high dimensional data will be projected to a GMM for low dimensional data, and the data distribution will not be nonlinearly warped nor linearly stretched. PCA also ensures that low dimensional data captures maximum total variance relative to the total variance in high dimensional data (alternatively, PCA projections minimize mean Euclidean squared error between high dimensional data and its approximation in the projection hyperplane). Therefore, we first applied PCA on $x$ to get the low dimensional representation of the data denoted by $\mathcal{Y}$. The parameters of the local Gaussians ($\mu_c$, $\Sigma_c$) are represented by $\Theta = [\theta_c]_{c=1}^{K}$, $\theta_c$, where $c$ ranges from 1 to $K$, with $K$ being the number of Gaussian components. The latent indicator is represented by $z$, and $\pi = [\pi_c]_{c=1}^{K}$ represents the weight sets for all local Gaussian components such that $\sum_{c=1}^{K} \pi_c = 1$. Therefore the probability density function for the GMM over $Y$ (with $Y$ representing all $\mathcal{Y}$ for the $x$ in the given dataset $X$) would be:

$$p(Y|\pi, \Theta) = \Pi_{n=1}^{N} \sum_{c=1}^{K} p(z_n|\pi) p(y_n|\theta_c)$$

Since we assume the distribution of $\mathcal{Y}$ is Gaussian or mixture of Gaussians, we then try to learn the parameters $\mu$ (mean) and $\Sigma$ (variance) for each of the local gaussians, using expectation-maximization algorithm.

**Cluster Interpretation**

*Comparing Clusters Across Participants Using Movie Clips (i.e., Trials)*

We compared cluster structure across participants by examining the extent to which clusters from different participants shared the same movie clips/trals. Specifically, we first assigned a single cluster label to each movie clip/trial based on the cluster for which it had the highest probability of belonging to. For every pair of clusters, across all participants and clusters, we then computed the percent of overlapping trials. Percent overlap was calculated as follows:

$$Percent\ overlap\ = \frac{N\ overlapping\ trials\ \times 2}{N\ trials\ in\ cluster\ A\ +\ N\ trials\ in\ cluster\ B} \times 100$$

Higher numbers indicate that trials cluster in similar ways, whereas lower numbers indicate that trials are clustering in unique ways. Lower numbers do not imply an absence of any similarity structure in trials within a given person, but rather they suggest that the similarity structure is not replicating across participants.

### *Assigning Meaning to Clusters*

We next examined whether the participant-specific clusters shared any variance with the ratings that were available in the data set (group-averaged emotion category ratings, affective/appraisal feature ratings, and semantic feature ratings, as well as session/run information). We used two different analyses.

First, we measured the information shared between the group-averaged emotion category ratings and the participant-level clusterings using normalized mutual information (NMI). We first computed Kullback-Leibler (KL) divergence $D_{KL}(P_{x|y} || P_x)$, where x represents the individual emotion category ratings and y represents the cluster assignments (with a Gaussian assumption for $P_x$ and $P_{x|y}$; Supplementary Figure 8). This measure tells us the ratio of entropy of the rating conditioned on the cluster label vs. the marginal entropy of the ratings (i.e., how much information is gained about the ratings by knowing the cluster membership). We then computed mutual information (MI) by taking the expectation of KL divergence with respect to the clusters. MI is formulated as H(X) - H(X|Y) or H(Y) - H(Y|X), where H represents the entropy of a random variable, x represents the individual emotion category ratings, and y represents the cluster assignments. Because entropy and MI can only assume positive values, one can bound MI using either H(X) or H(Y). We normalized MI (such that NMI ∈ [0,1]) by dividing by H(Y), the entropy over the cluster assignment. The resultant value gives us the coefficient of uncertainty. NMI calculations for session and run utilized the built-in scikit-learn NMI function because these variables were from discrete distributions, rather than continuous Gaussian distributions. A similar process was followed to assess the degree of relation between the group-averaged affective/appraisal feature ratings, the group-averages semantic feature ratings, and the design (session/run) information.

Using a second method, we qualitatively examined the highest-rated emotion category per trial amongst each cluster. We assigned a single emotion category label to each trial, and then plotted the distribution of these labels amongst each participant's clusters. A similar analysis was performed for the group-averaged affective/appraisal feature ratings, the group-averages semantic feature ratings, and the design (session/run) information.

**Acknowledgments**

The authors thank Dr. Dana Brooks, Dr. Jordan Theriault, and Dr. Danlei Chen for meaningful discussions and insight on our modeling and analyses. This work was supported by grants from the National Science Foundation (BCS 1947972), the National Institutes of Health (R01 AG071173, R01 MH109464, and R01 MH113234), the U.S. Army Research Institute for the Behavioral and Social Sciences (W911NF-16-1-019), and the Unlikely Collaborators Foundation. The views, opinions, and/or findings contained in this manuscript are those of the authors and shall not be construed as an official Department of the Army position, policy, or decision, unless so designated by other documents, nor do they necessarily reflect the views of the Unlikely Collaborators Foundation.

**Data Availability**

The data used in this study are available from the original publications (Cowen & Keltner, 2017; Horikawa et al., 2020).

**Author Contributions**

**Figure Legends**

**Figure 1.** Plots of mean BIC across 100 iterations used to determine the optimal number of components to use in each participant's GMM. X-axes represent the number of components, and y-axes represent the BIC values. The optimal number of components, shown in the bottom right table, is the value that minimizes the BIC.

**Figure 2.** Percent of overlapping trials across all pairs of clusters. Axes represent participant and cluster names (e.g., P1_C01 = participant 1, cluster 1). Black lines separate clusters for each participant. Percent overlap ranges from 0 (light yellow) to 100 (dark red).

**Figure 3.** Plots of normalized mutual information between each participant's clustering and group-averaged emotion category ratings. X-axes represent the participant number, and y-axes represent the emotion category label. a) Results shown across the full scale from 0 (light yellow) to 1 (dark red). b) The same results shown across a restricted scale from 0 to 0.1.

**Figure 4.** Distribution of trials in each cluster for one example participant (Sub-01), plotted based on different potential sources of variance. Plots depict (a) the highest rated emotion category for that trial, (b) the highest rated affective/appraisal features for that trial, (c) the highest rated semantic features for that trial, (d) the corresponding session for that trial, (e) the corresponding run for that trial. Each bar represents a single cluster. Trials are colored based on cluster. Contempt, envy, and guilt were not rated highest for any video, and thus are not shown in (a).

EXPLORING THEORY-LADEN OBSERVATIONS 19

**Figures**

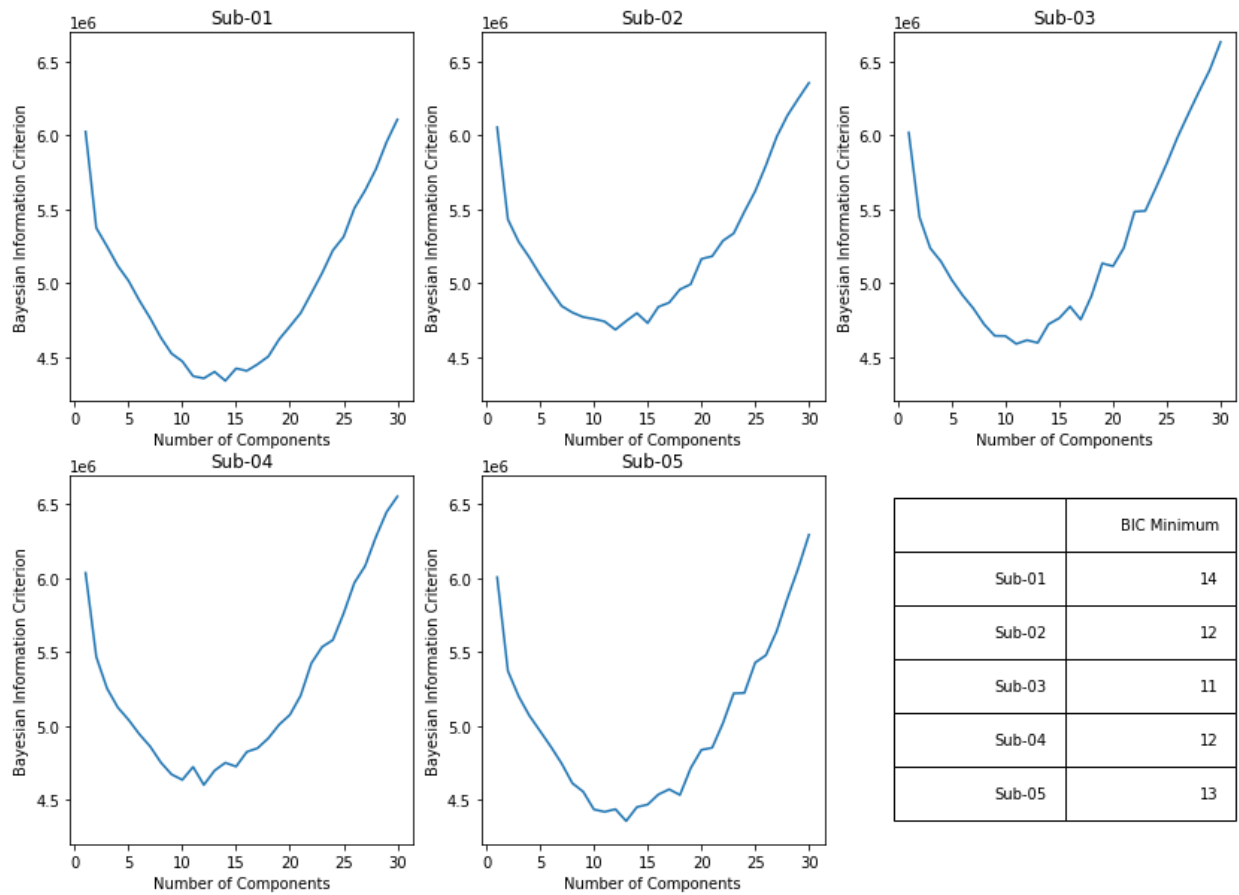

**Figure 1.** Plots of mean BIC across 100 iterations used to determine the optimal number of components to use in each participant's GMM. X-axes represent the number of components, and y-axes represent the BIC values. The optimal number of components, shown in the bottom right table, is the value that minimizes the BIC.

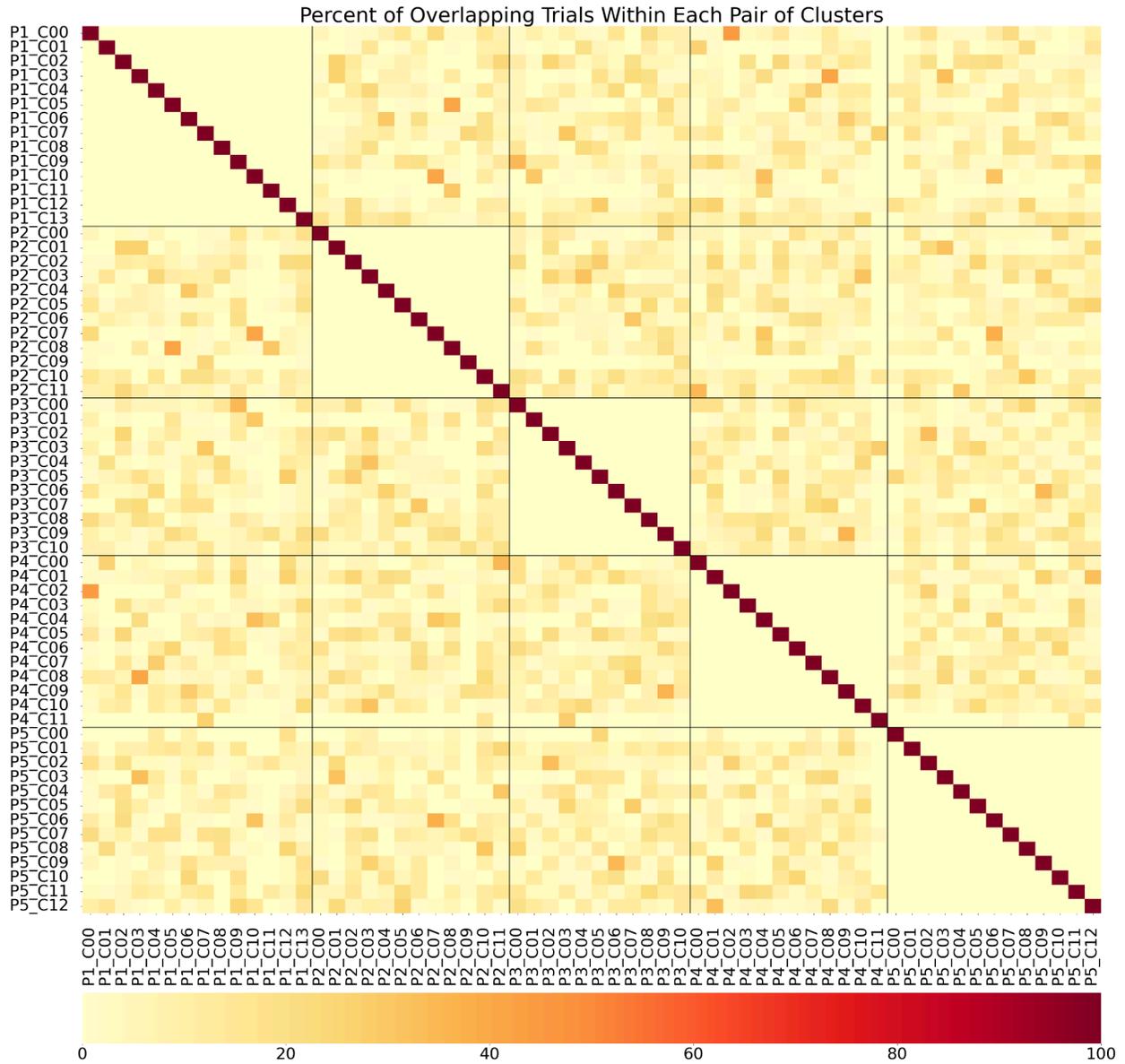

**Figure 2.** Percent of overlapping trials across all pairs of clusters. Axes represent participant and cluster names (e.g., P1_C01 = participant 1, cluster 1). Black lines separate clusters for each participant. Percent overlap ranges from 0 (light yellow) to 100 (dark red).

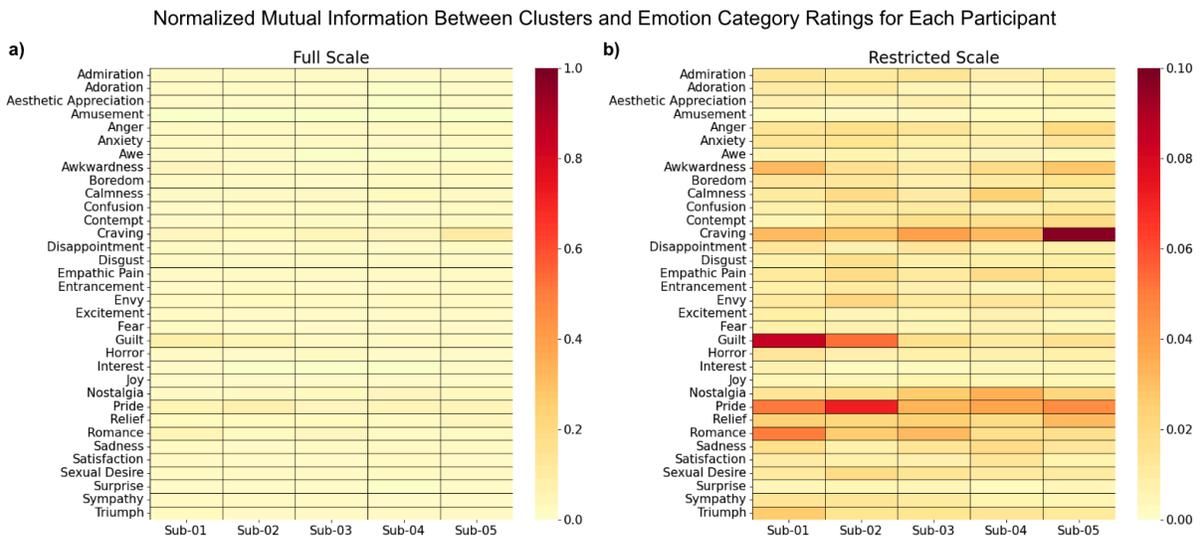

**Figure 3.** Plots of normalized mutual information between each participant's clustering and group-averaged emotion category ratings. X-axes represent the participant number, and y-axes represent the emotion category label. a) Results shown across the full scale from 0 (light yellow) to 1 (dark red). b) The same results shown across a restricted scale from 0 to 0.1.

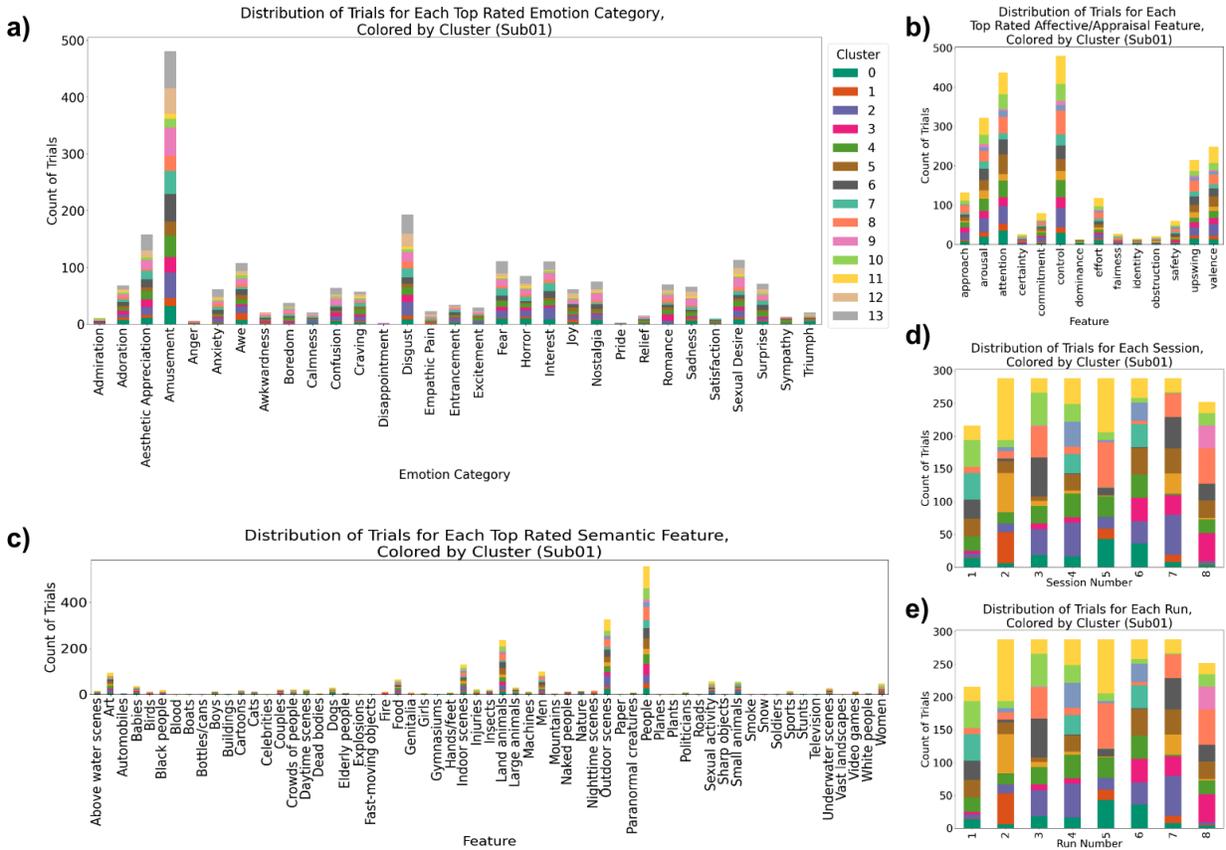

**Figure 4.** Distribution of trials in each cluster for one example participant (Sub-01), plotted based on different potential sources of variance. Plots depict (a) the highest rated emotion category for that trial, (b) the highest rated affective/appraisal features for that trial, (c) the highest rated semantic features for that trial, (d) the corresponding session for that trial, (e) the corresponding run for that trial. Each bar represents a single cluster. Trials are colored based on cluster. Contempt, envy, and guilt were not rated highest for any video, and thus are not shown in (a).

## Supplementary Materials

**Cosine Similarity Between Cluster Means**

To compare clusters across participants, we applied an inverse transformation on the mean of each cluster to transform it back into the original data space (i.e., low dimensional PCA space to high dimensional fMRI space; (Bishop, 2006)), and then computed cosine similarity between every pair of cluster means across all participants. The resultant cosine similarity value provides an index of how similar clusters were in terms of their mean BOLD signal, where a value of 1 suggests highly similar means, a value of 0 suggests orthogonal means, and -1 suggests opposite means. Mean cosine similarity values were close to zero (mean cosine similarity between pairs of clusters within-participants = -0.0052, between-participants = 0.0122; Figures 8-11), suggesting individual participants BOLD data clustered together in variable ways.

We also compared cluster means within two *a priori* defined regions: primary visual cortex (V1) and ventromedial prefrontal cortex (vmPFC). We parcellated the brain into 50 regions using the parcellation reported in (Chang et al., 2021), and then extracted the means for each cluster for parcels corresponding to the regions of interest. Cosine similarity was then computed on these means using the same approach as was done on the whole-brain data. The mean similarity of V1 (within-participant mean cosine similarity = 0.0601; between-participants mean cosine similarity= 0.0533) was higher than that of vmPFC (within-participant mean cosine similarity = 0.0012; between-participants mean cosine similarity= 0.0055). Despite overall low similarity in both regions, the similarity in V1 was approximately 50 times higher within-participants than similarity in vmPFC, and approximately 10 times higher between-participants. These findings suggest that lower-level visual features represented in V1 are more shared across clusters, but the higher-level, compressed summaries of these features represented in vmPFC are more variable. In other words, while people are viewing similar things, the meaning of these things is experienced in more variable ways.

**PCA Tests**

When the dimensionality of the data is much larger than the number of the samples, there are several problems that may arise with the application of Principal Component Analysis (PCA) Specifically, there may be an upward bias in the eigenvalues, as well as inconsistency in eigenvectors (see (Johnstone & Paul, 2018) for a detailed explanation of phenomena that might arise in this situation of mismatched dimensionality). We explored the potential for these problems in our data through various tests. Because our implementation of PCA was largely based on the singular value decomposition (SVD) of the covariance matrix, we explored the elements of the SVD of the covariance matrix (i.e., eigenvalues and eigenvectors) through the following tests:

1. First, we examined the spread of the eigenvalues. The spread of the eigenvalues revealed how much of the variance in the original covariance matrix was explained by each of the eigenvectors/principal components. With high-dimensional data we would ideally observe that only a few components are required to capture most of the variance, meaning that the first few eigenvalues should be very large compared to the trailing eigenvalues. If the true dominant direction of variation in the data is being represented by the principal components, then the spread of eigenvalues across randomly sampled subsamples of the data should all look alike (i.e., all skewed). To test this, we selected random subsamples of one participant's data (n= 200, 500, 1000, 1500, and 2000) and visualized the spread (Supplementary Figure 13). Two observations can be drawn from the spread of our data: (1) the first few eigenvalues are dominant compared to the remaining eigenvalues, and (2) as the number of samples increased, the spread of the eigenvalues did not change. These observations suggest that a low rank approximation of the true covariance matrix (belonging to the population) would retain most variance in the data.

2. Second, we examined the consistency of the eigenvectors across different iterations of increasing sample sizes. If the direction of variation in the high-dimensional data is consistent, we would expect to see strong similarity between the top eigenvectors (sorted in order of explained variance) representing the covariance matrices achieved on different randomly selected subsamples of the data. To test this, we compared the cosine distances between the top five principal components of random subsamples of one participant's data (n = 200, 500, 1000, 1500, and 2000) across 10 iterations of random sampling for each sample size (Supplementary Figure 14). As sample size increased, the cosine similarity between principal components tended towards one. This finding suggests that the dominant eigenvectors are consistent across different random samples. Supplementary Figure 15 depicts cosine distance between the top most dominant principal component across different sample sizes.

3. Third, we examined reconstruction loss to evaluate how well the data could be recovered in the original dimensions with respect to the lower dimensional projection. The lower the reconstruction loss is, the better the original data was able to be recovered. Reconstruction loss also provides information on whether the low dimensional encoding is able to capture sufficient variation, in the original dimension, to be able to reconstruct with less error. To test this, we compared the reconstruction loss of a test set (n = 220 samples) when principal components were estimated from a training set containing varying subsamples of one participant's data (n = 100 to 2000 samples; Supplementary Figure 16). We observed a decrease in reconstruction loss of the left-out test set as the number of samples in the training set increased. When the sample size increased to greater than 1000, the reconstruction loss has already converged to a very small number, suggesting that the reconstruction loss at our sample size of 2196 data points was very low.

**Supplementary Figures**

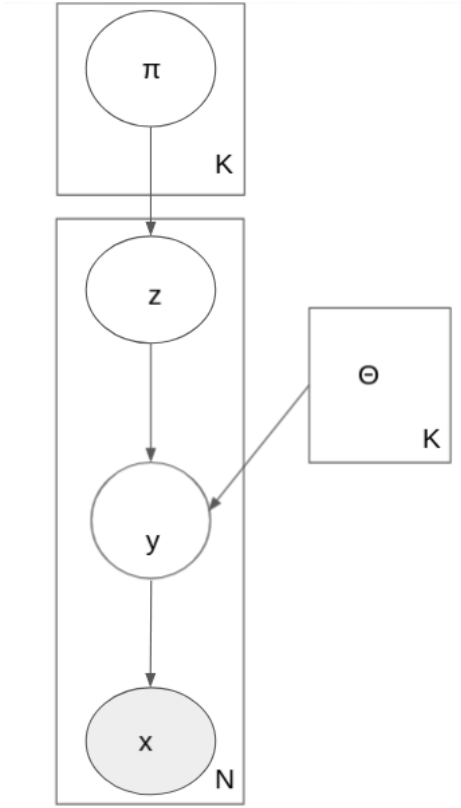

**Supplementary Figure 1.** Probabilistic Graphical Model of the PCA-GMM analysis.

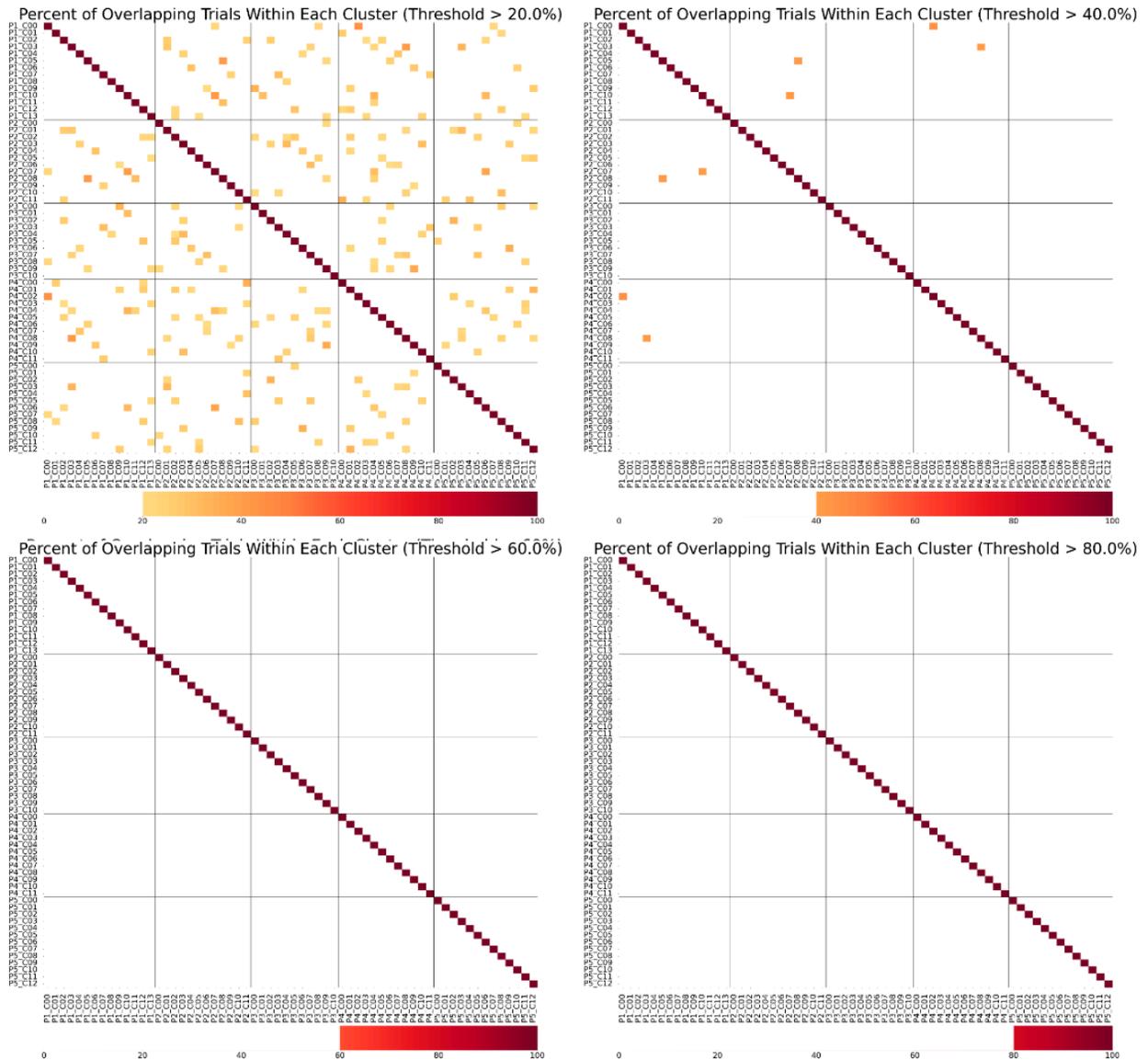

**Supplementary Figure 2.** Percent of overlapping trials across all pairs of clusters at four thresholds: overlap > 20% (top left), 40% (top right), 60% (bottom left), and 80% (bottom right). Axes represent participant and cluster names (e.g., P1_C01 = participant 1, cluster 1). Black lines separate clusters for each participant. White values indicate overlap lower than the threshold.

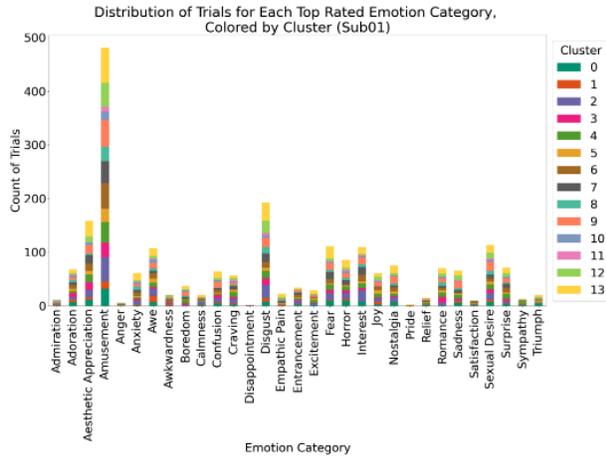
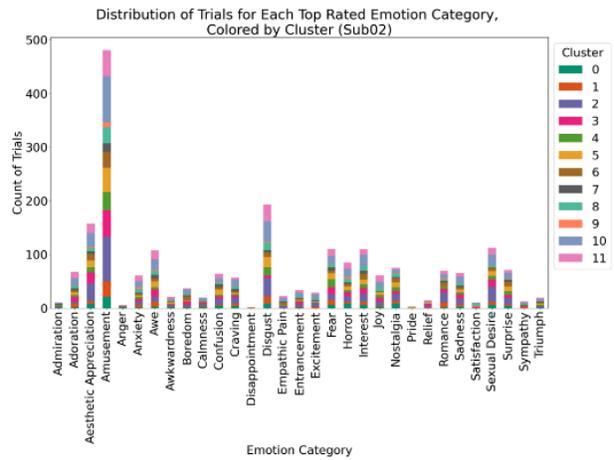
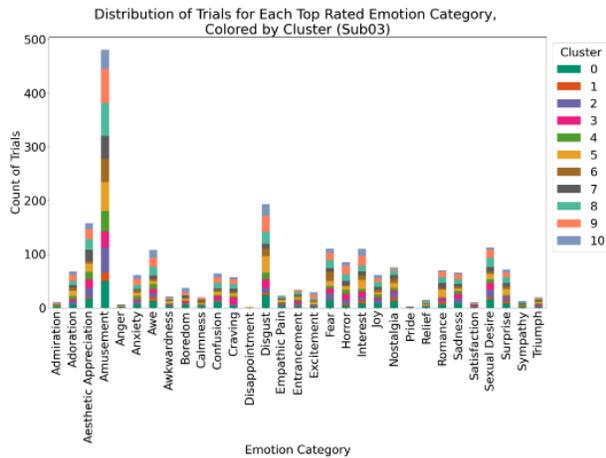
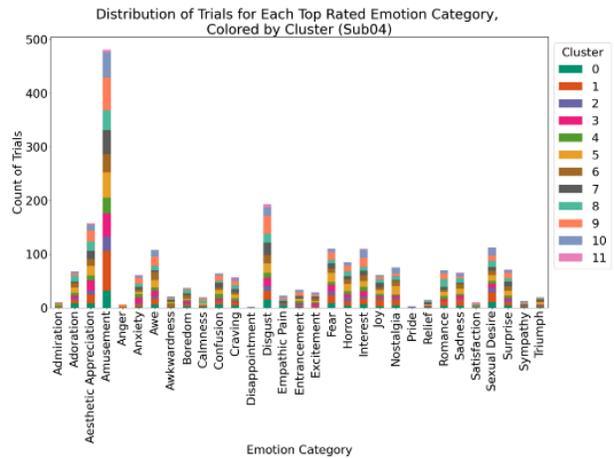
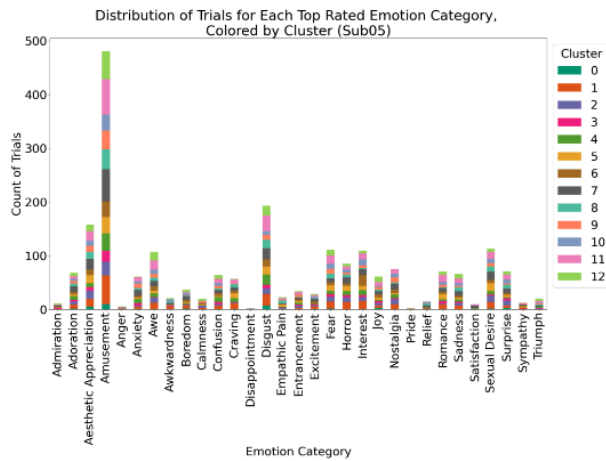

**Supplementary Figure 3.** Distribution of trials per cluster for each top-rated emotion category in each participant's clusters. Each bar represents a single emotion category. Trials are colored based on cluster. Contempt, envy, and guilt were not rated highest for any video, and thus are not shown.

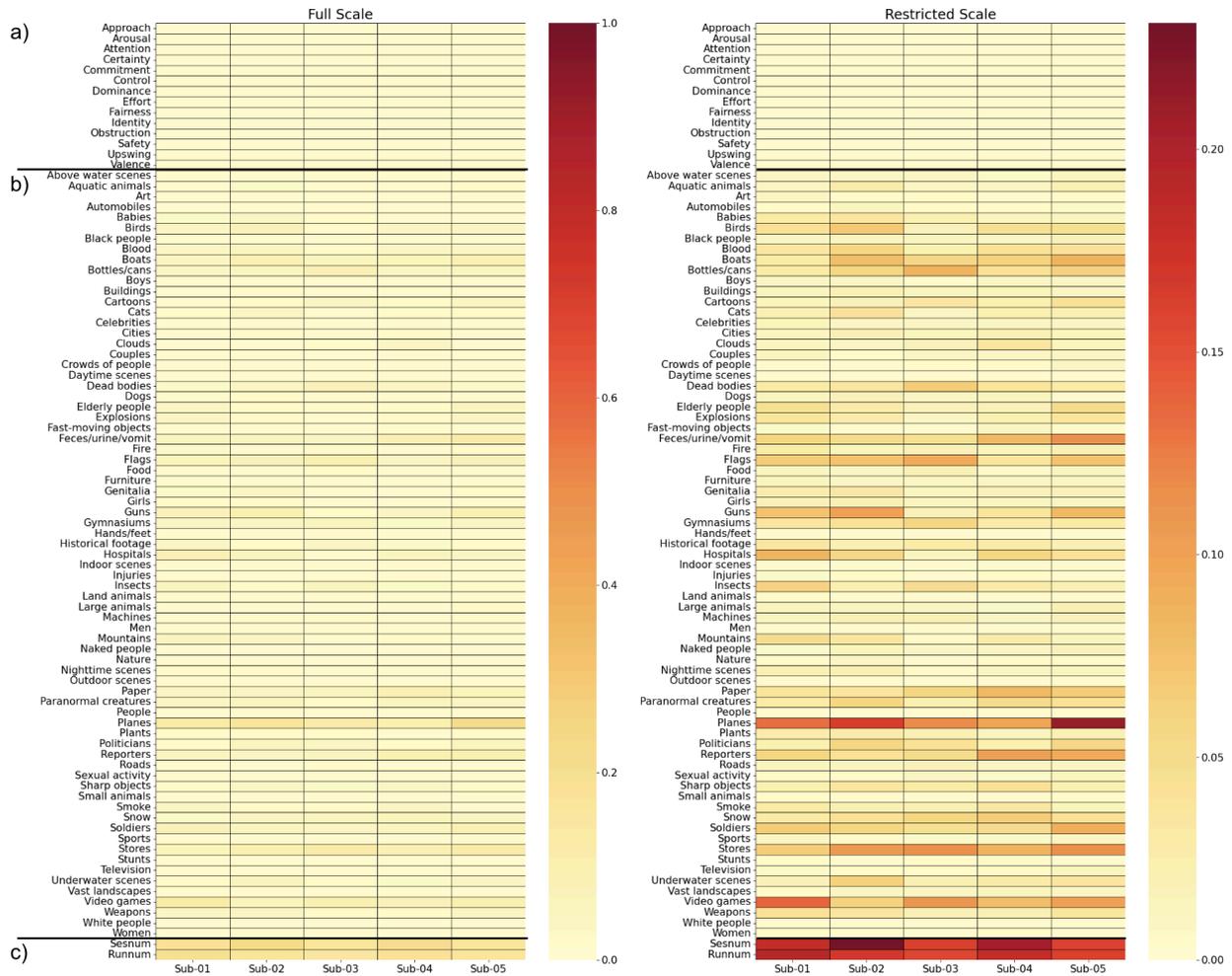

**Supplementary Figure 4.** Plots of normalized mutual information between each participant's clustering and other sources of variance. Specifically, plots show group-averaged ratings of (a) affective features and (b) semantic features, as well as (c) session/run information. X-axes represent the participant number, and y-axes represent the feature variable. Left plots show NMI across the full scale from 0 to 1, right plots show the same results across a restricted scale.

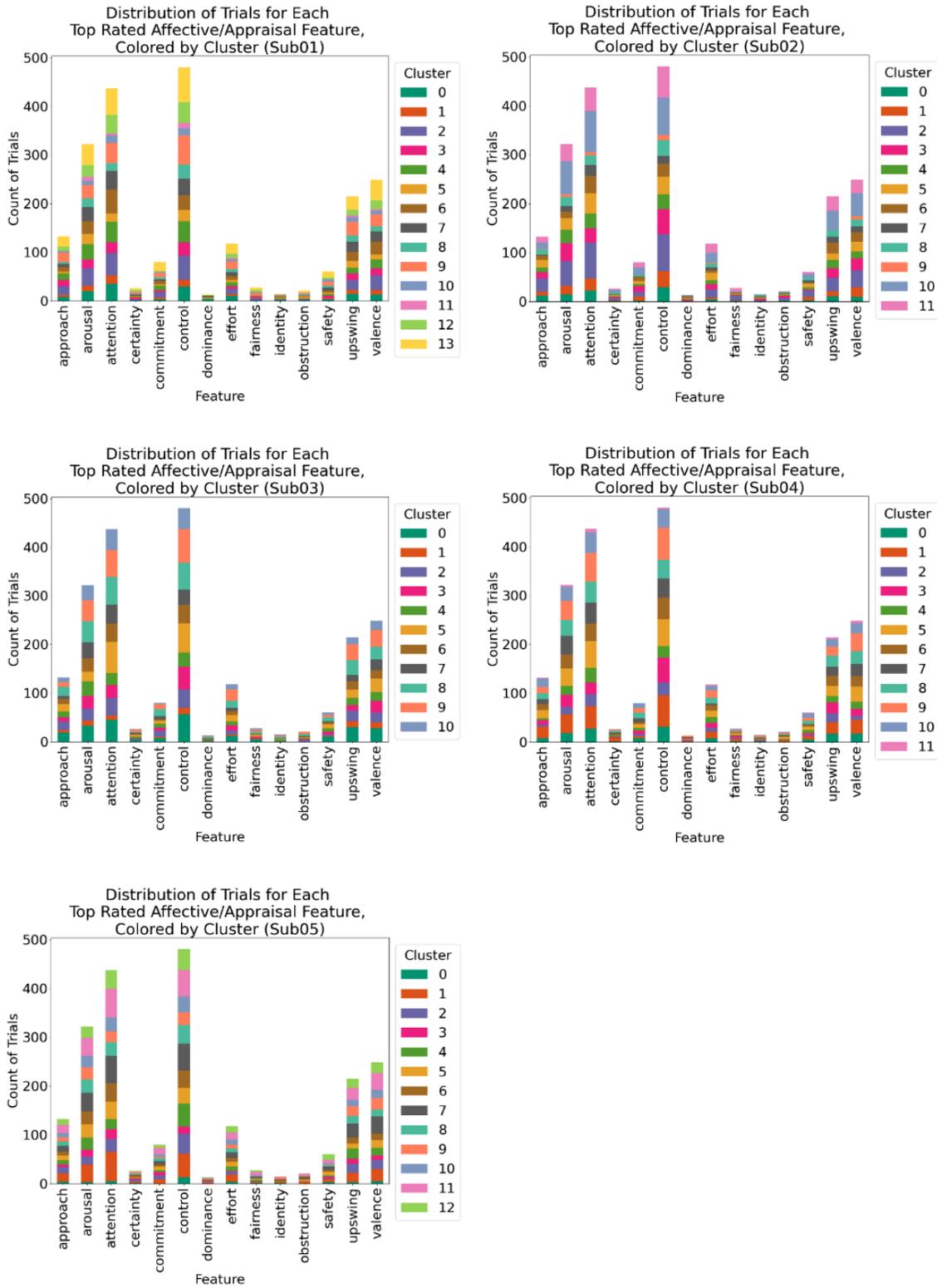

**Supplementary Figure 5.** Distribution of trials per cluster for each top-rated affective/appraisal feature in each participant's clusters. Each bar represents a single feature. Trials are colored based on cluster.

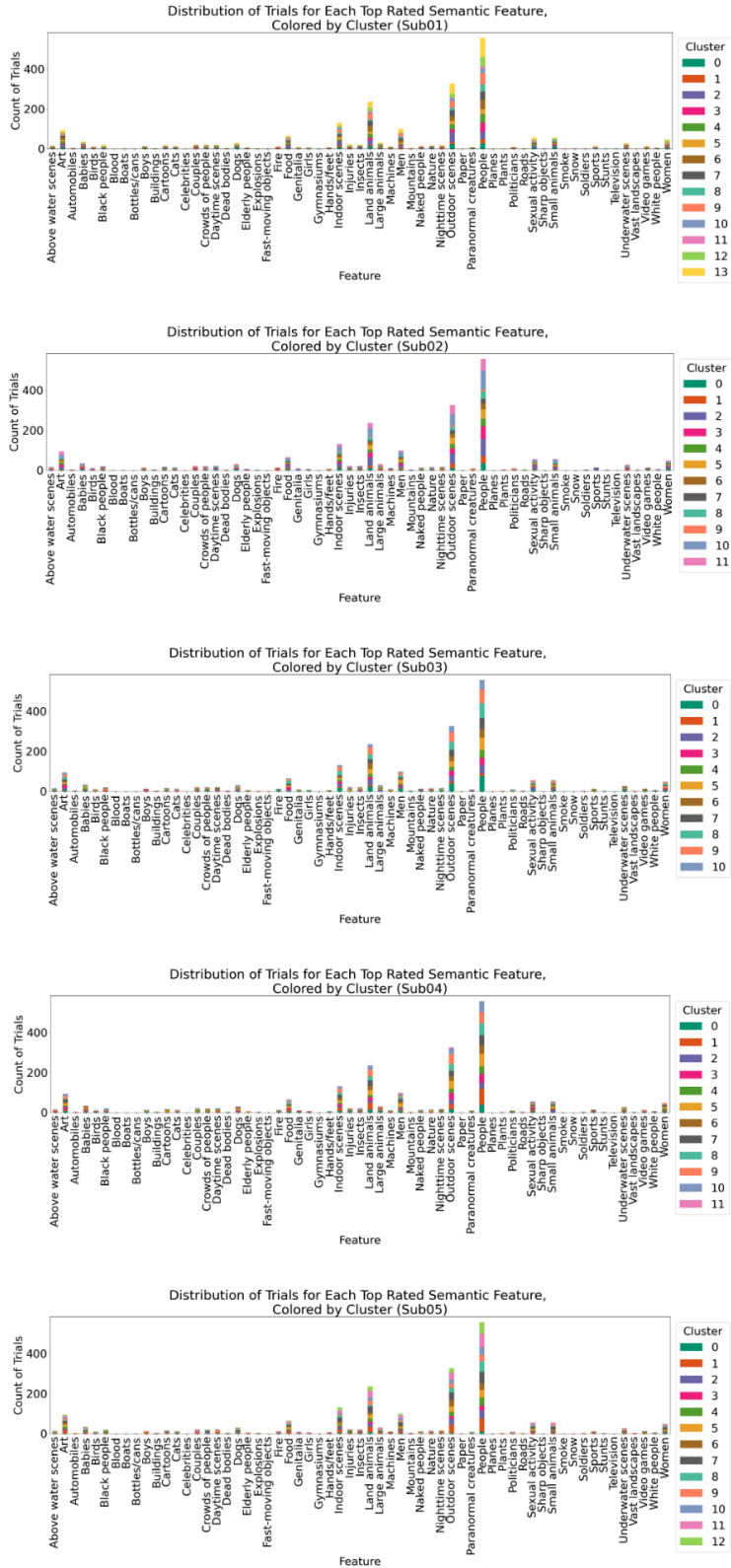

**Supplementary Figure 6.** Distribution of trials per cluster for each top-rated semantic feature in each participant's clusters. Each bar represents a single feature. Trials are colored based on cluster.

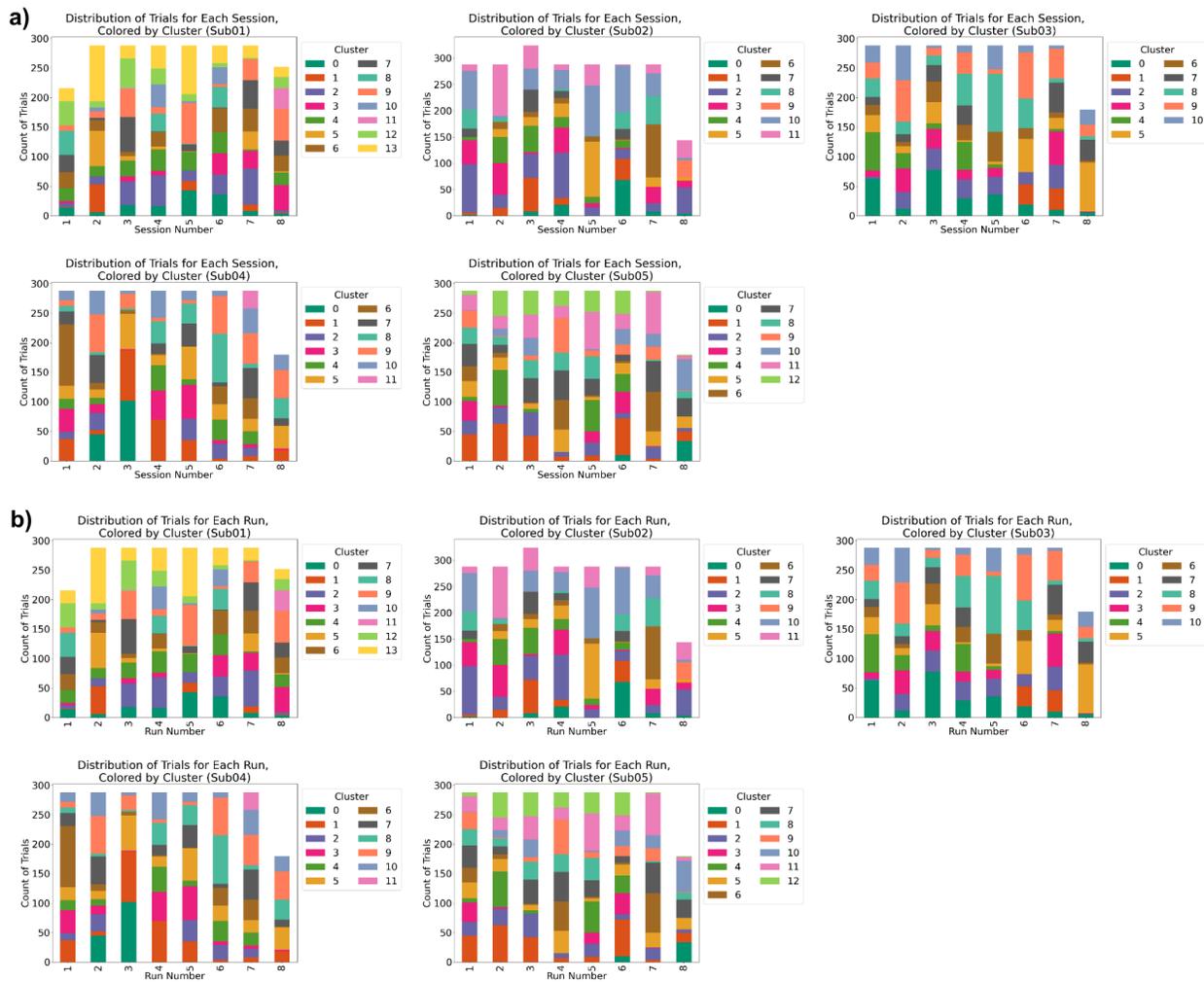

**Supplementary Figure 7.** Distribution of trials per cluster for each session (a) and run (b) of trials in each participant's clusters. Each bar represents a single cluster. Trials are colored based on cluster.

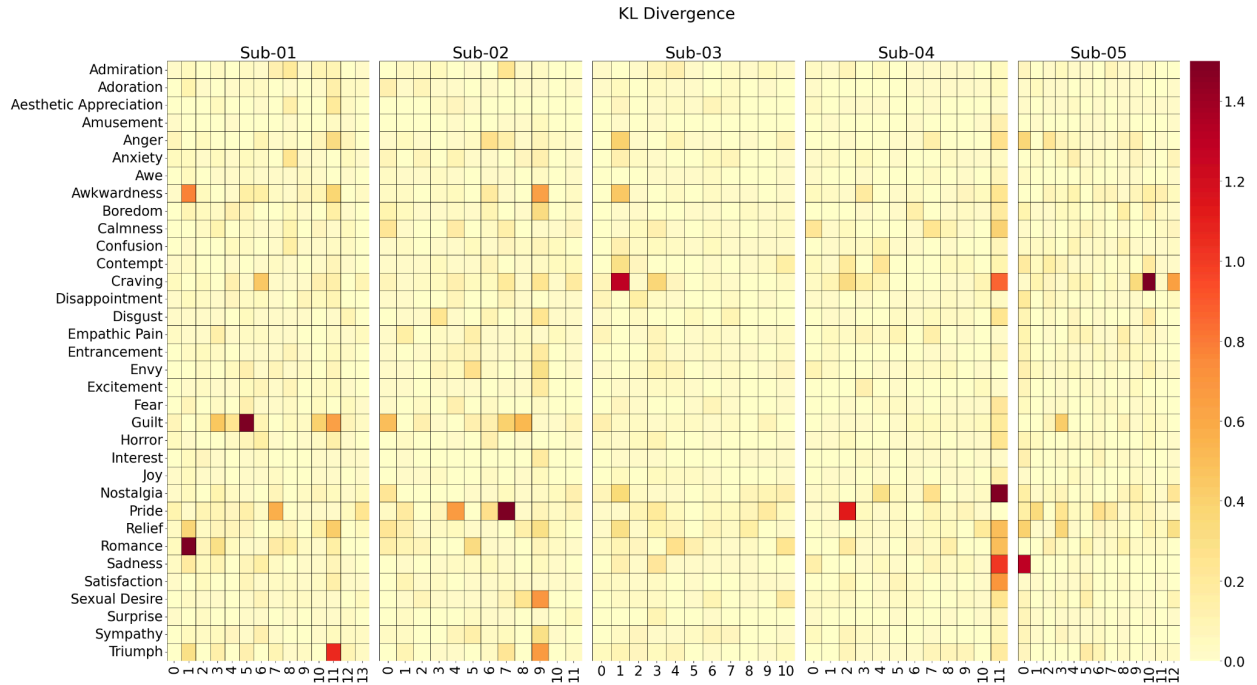

**Supplementary Figure 8.** Plots of KL Divergence shared between $(P_{x|y} || P_x)$ for each participant, where x represents the individual emotion category ratings and y represents the cluster assignments. X-axes represent the cluster number, and y-axes represent the emotion category label.

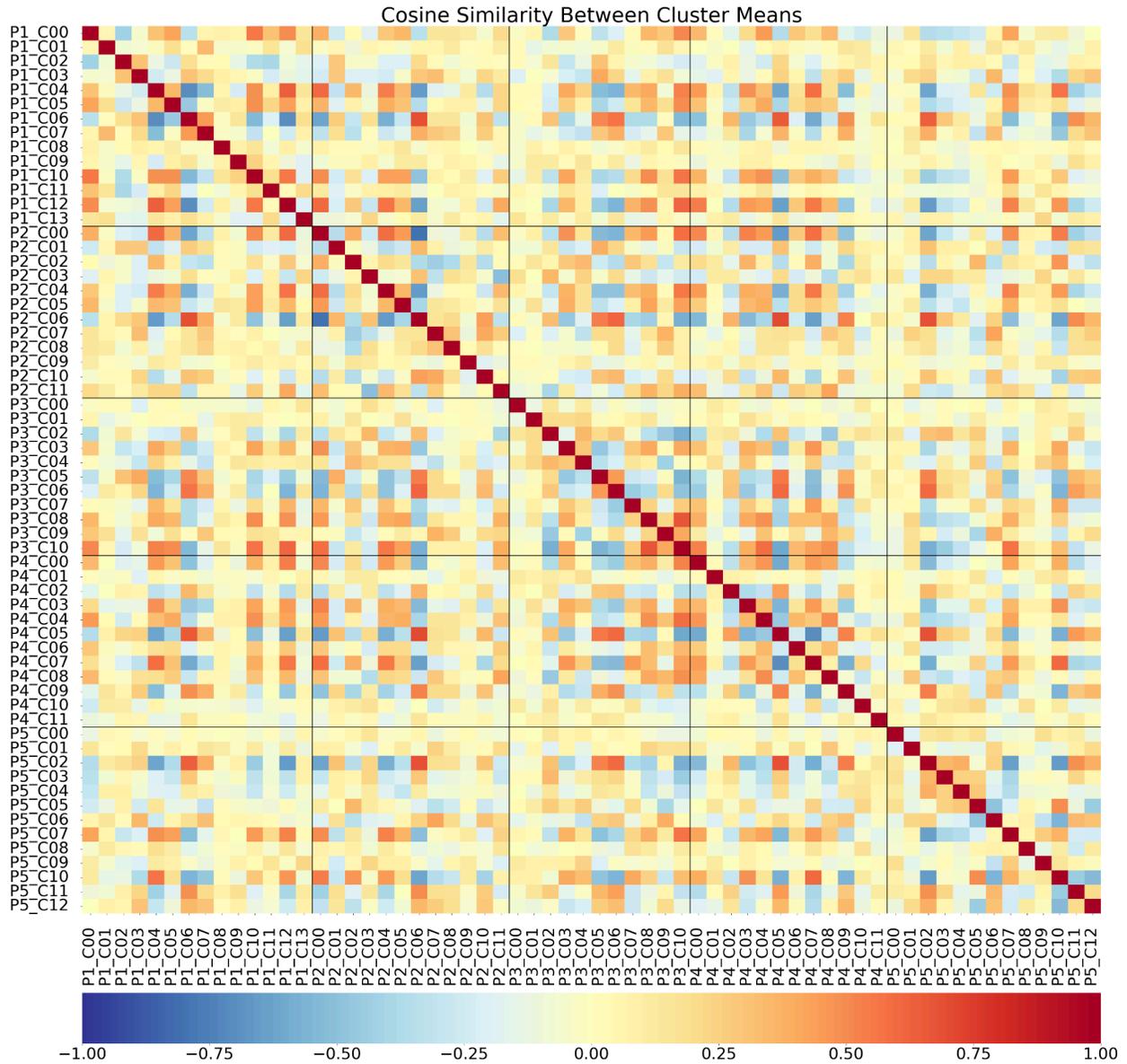

**Supplementary Figure 9.** Cosine similarity between cluster means across all pairs of clusters. Axes represent participant and cluster numbers (e.g., P1_C01 = participant 1, cluster 1). Black lines separate clusters for each participant. Cosine similarity ranges from -1 (dark blue) to 1 (dark red). A value close to 1 suggests highly similar mean vectors between the two clusters, a value close to 0 suggests distinct cluster means, and a value close to -1 suggests the mean vectors are opposite. The distribution of all cosine similarity values is shown in Supplementary Figure 3.

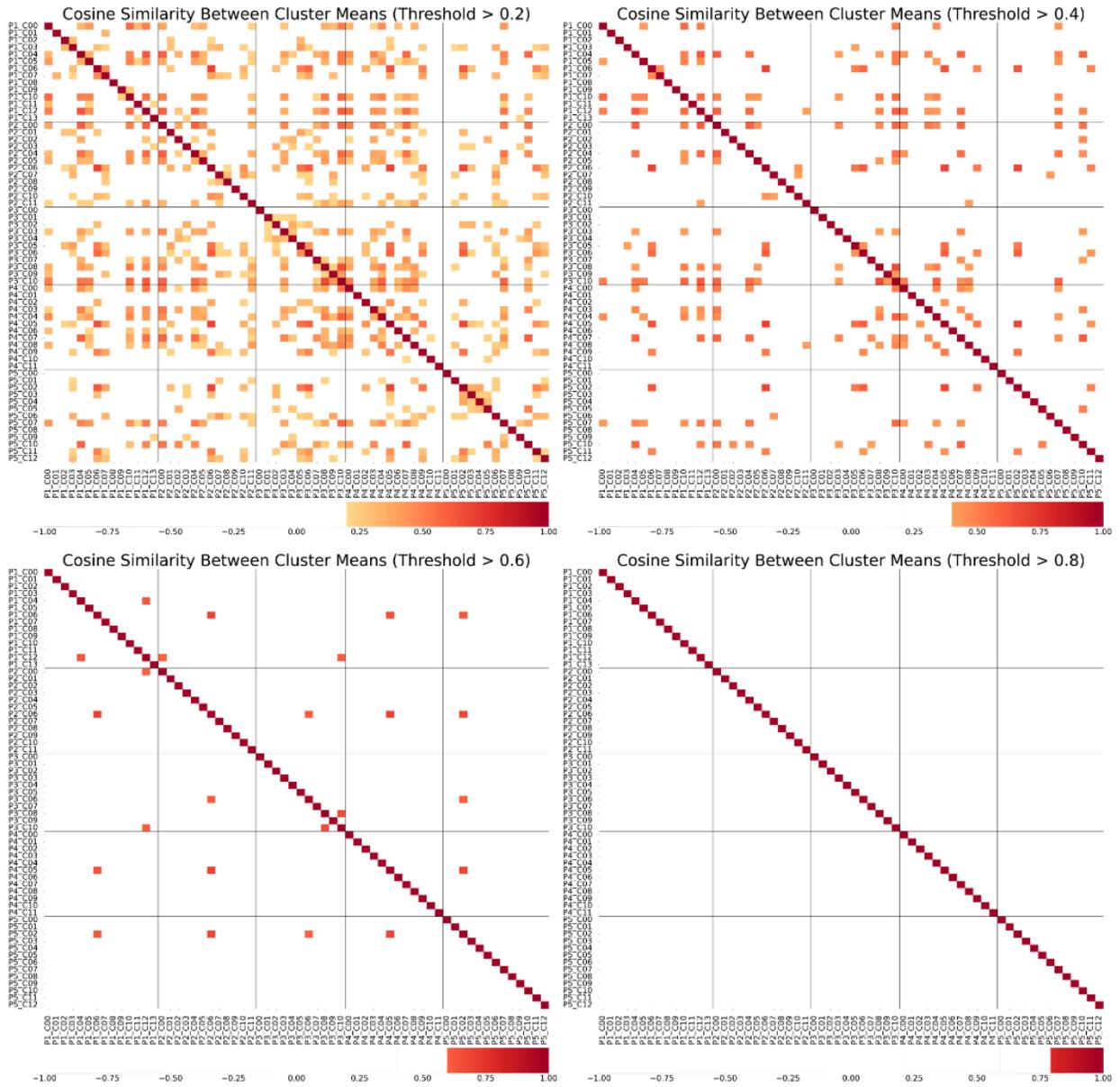

**Supplementary Figure 10.** Cosine similarity between cluster means across all pairs of clusters at four thresholds: cosine similarity > 0.2 (top left), 0.4 (top right), 0.6 (bottom left), and 0.8 (bottom right). Axes represent participant and cluster numbers (e.g., P1_C01 = participant 1, cluster 1). Black lines separate clusters for each participant. White values indicate similarity lower than the threshold.

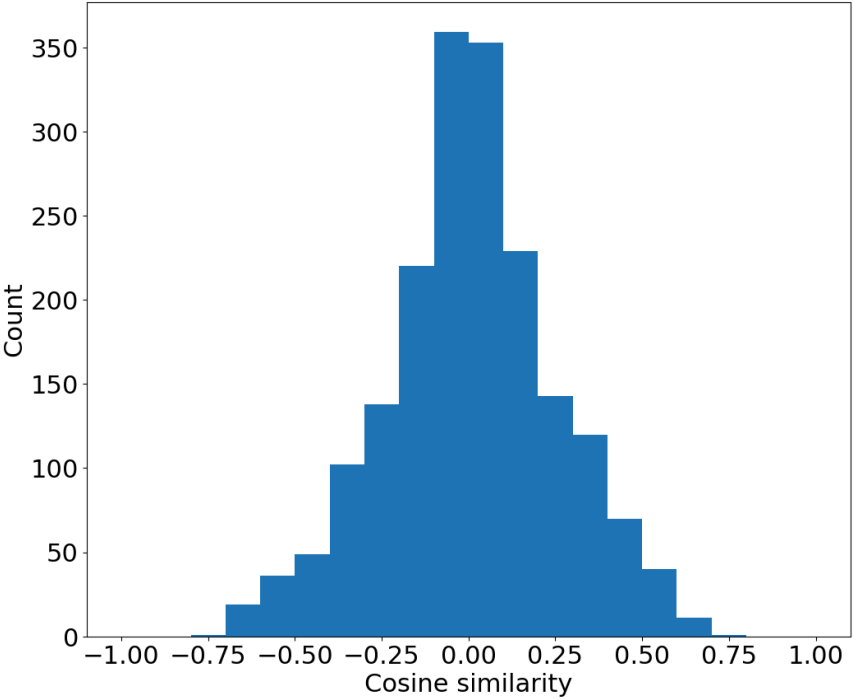

**Supplementary Figure 11.** Distribution of cosine similarity values between cluster means across all pairs of clusters.

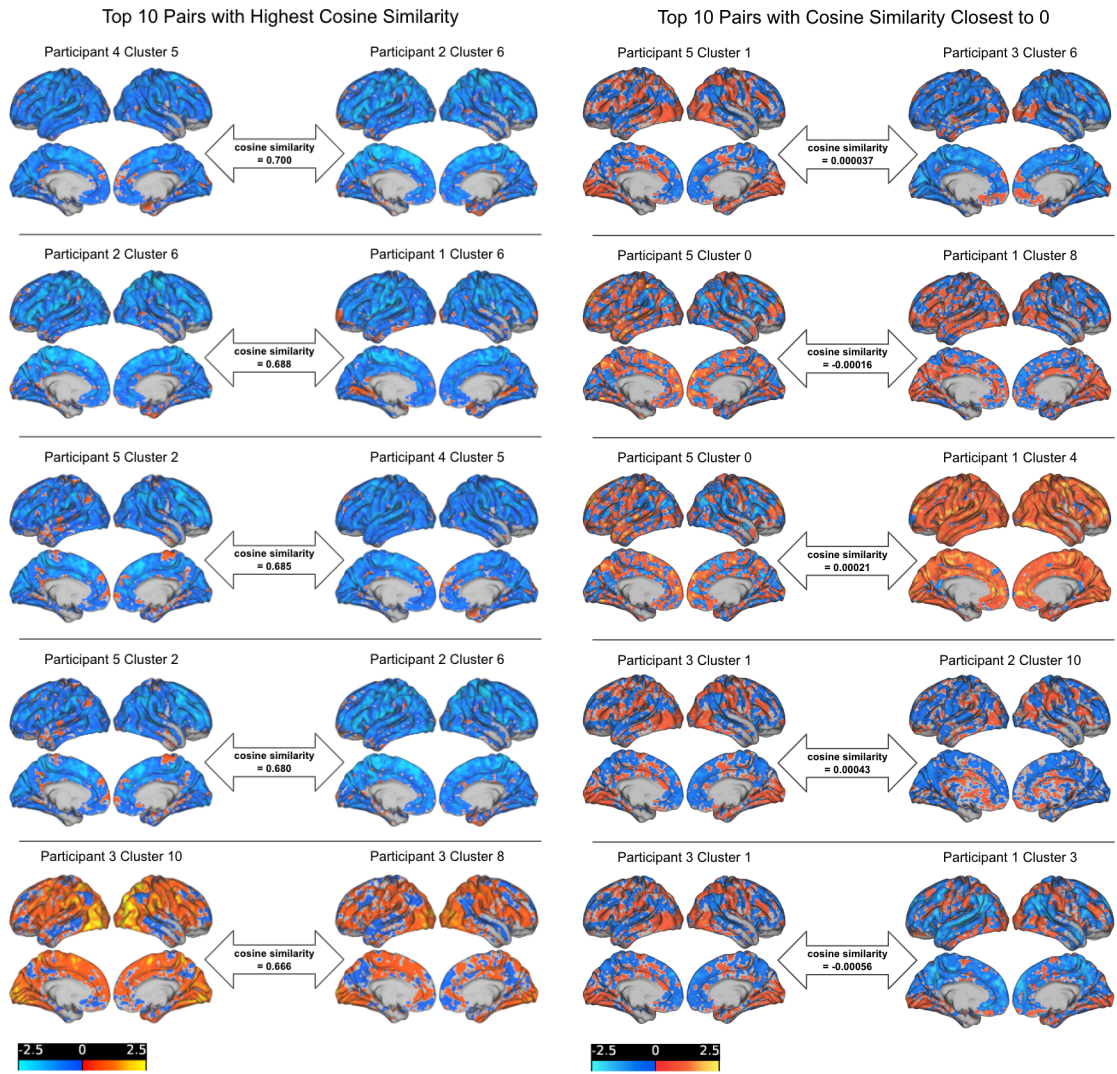

**Supplementary Figure 12.** Visualization of the cluster means (i.e., mean BOLD signal) projected back into high-dimensional fMRI space on the cortex for the top five pairs with the highest cosine similarity (left) and the five pairs with the cosine similarity closest to zero (right).

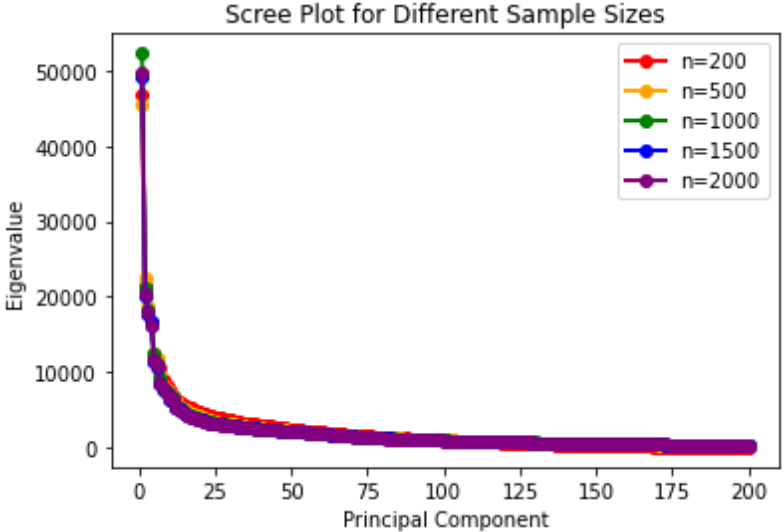

**Supplementary Figure 13.** Scree plot of eigenvalues for one representative participant (Participant 01). Each colored line represents different sample sizes when randomly selecting 200, 500, 1000, 1500, and 2000 samples from our data.

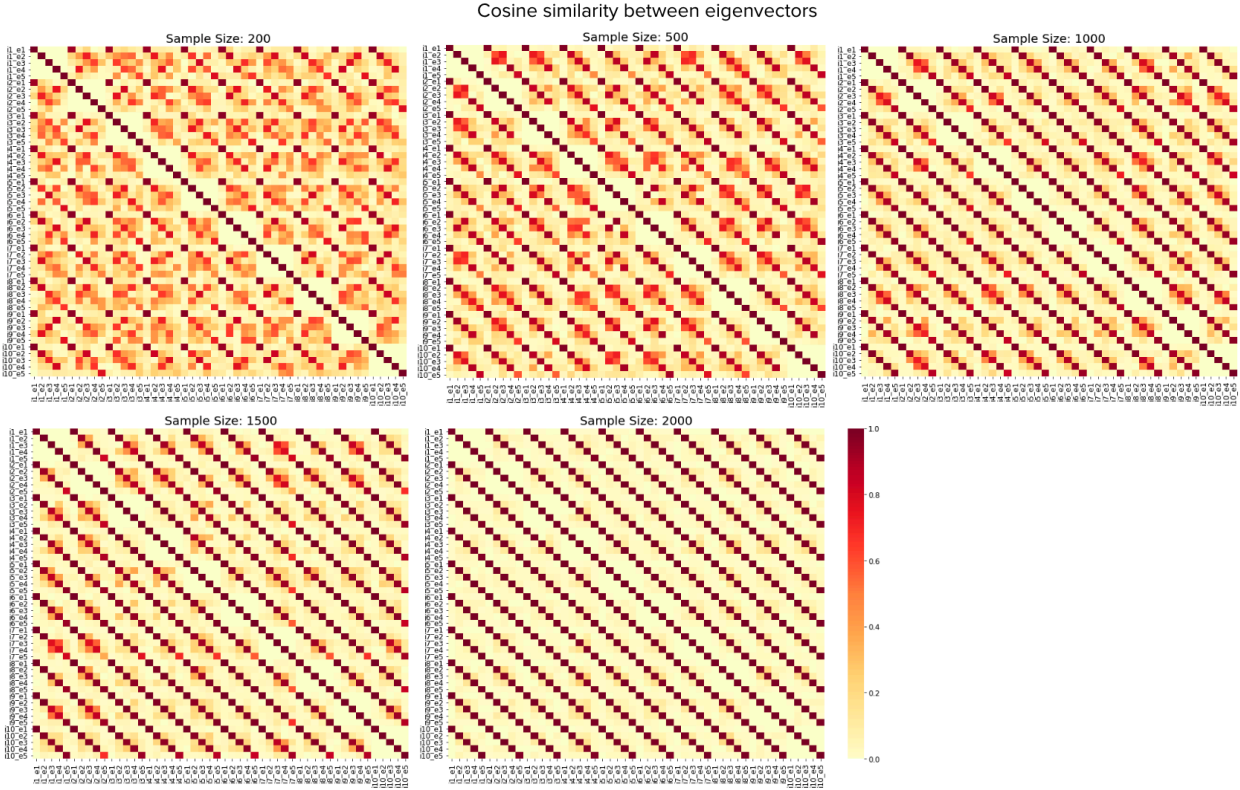

**Supplementary Figure 14.** Cosine similarity for one representative participant (Participant01) between the first five eigenvectors across 10 iterations. Each plot depicts a cosine similarity matrix for different sample sizes when randomly selecting 200, 500, 1000, 1500, and 2000 samples from our data. Axis labels reflect iteration and eigenvector ('i1_e1' refers to iteration 1, eigenvector 1).

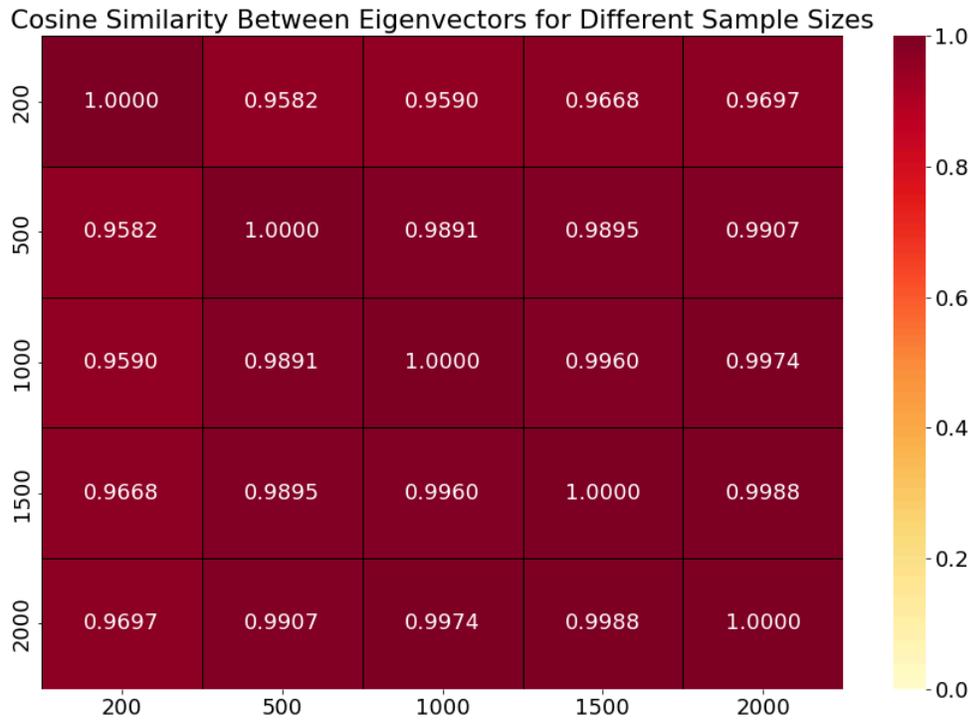

**Supplementary Figure 15.** Cosine similarity between the eigenvectors corresponding to the first principal component for 200, 500, 1000, 1500, and 2000 random samples of one representative participant's data (Participant01).

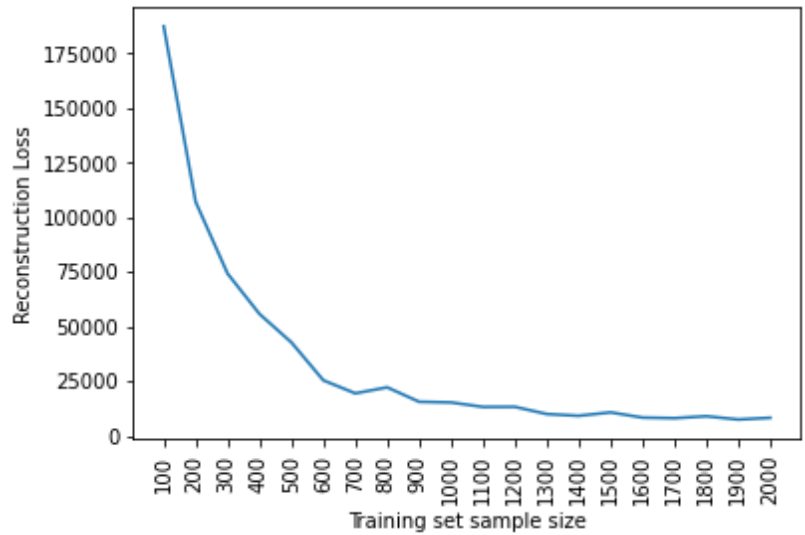

**Supplementary Figure 16.** Reconstruction loss for one representative participant's data (Participant01) to test out-of-sample robustness along different sizes of randomly sampled training sets. Reconstruction was calculated on a left out test set of 220 samples.